\documentclass{article}
\usepackage{caption}
\usepackage{forest}
\usepackage{multicol}
\usepackage{subcaption}
\usepackage[a4paper, total={6in, 10in}]{geometry}
\usepackage[utf8]{inputenc}
\usepackage{graphicx} % pdf to figure
\usepackage{indentfirst} % indent after section
\usepackage{amsmath,amsfonts,amssymb}% 数学符号
\usepackage{tabularx}% 公式+符号说明
\usepackage{amsmath}
\usepackage{mathtools}
\newcommand{\veryshortarrow}[1][3pt]{\mathrel{%
   \hbox{\rule[\dimexpr\fontdimen22\textfont2-.2pt\relax]{#1}{.4pt}}%
   \mkern-4mu\hbox{\usefont{U}{lasy}{m}{n}\symbol{41}}}}

\usepackage{xcolor}% 颜色
\newenvironment{conditions*}
  {\par\vspace{\abovedisplayskip}\noindent
   \tabularx{\columnwidth}{>{$}l<{$} @{${}:{}$} >{\raggedright\arraybackslash}X}}
  {\endtabularx\par\vspace{\belowdisplayskip}}
\usepackage{booktabs,dcolumn,caption}

\usepackage{bm}
\usepackage{float}
\usepackage{amsfonts}
\usepackage{array}
\usepackage[ruled,vlined]{algorithm2e}
\usepackage{hyperref}
\newcommand{\specialcell}[2][c]{%
  \begin{tabular}[#1]{@{}c@{}}#2\end{tabular}}

\title{A Comprehensive Introduction of Visual-Inertial Navigation}
\author{Yangyang NING}
\date{March 2022}

\begin{document}
\maketitle

\section*{Abstract}
In this article, a tutorial introduction to visual-inertial navigation(VIN) is presented. Visual and inertial perception are two complementary sensing modalities. Cameras and inertial measurement units (IMU) are the corresponding sensors for these two modalities. The low cost and light weight of camera-IMU sensor combinations make them ubiquitous in robotic navigation. Visual-inertial Navigation is a state estimation problem, that estimates the ego-motion and local environment of the sensor platform. This paper presents visual-inertial navigation in the classical state estimation framework, first illustrating the estimation problem in terms of state variables and system models, including related quantities representations (Parameterizations), IMU dynamic and camera measurement models, and corresponding general probabilistic graphical models (Factor Graph). Secondly, we investigate the existing model-based estimation methodologies, these involve filter-based and optimization-based frameworks and related on-manifold operations. We also discuss the calibration of some relevant  parameters, also initialization of state of interest in optimization-based frameworks. Then the evaluation and improvement of VIN in terms of accuracy, efficiency, and robustness are discussed. Finally, we briefly mention the recent development of learning-based methods that may become alternatives to traditional model-based methods.

%are able to indirectly model dynamic of IMU and replace part of the process of motion estimation
\section{Introduction}
% Brief Intro
Environment and ego-motion perception is critically important for navigation. Motion perception in biology offers a more intuitive perspective, human use various sensory modalities to feel self-motion and surroundings, especially combining motion and balance from the inner ear, joint position, and visual information from the eyes to obtain a virtual sense of movement, called kinesthesia\cite{RN310}. Instead, motion perception of robots is provided through the use of navigation and positioning techniques including dead reckoning methods (using inertial sensors,  pedometer, wheel encoder, magnetometer, and gyrocompass, etc.) and position fixing methods (using global navigation satellite system (GNSS), ultra-wideband (UWB), acoustic ranging, lidar, and cameras, etc.)\cite{RN12}.
In this work, we focus on two types of sensory modalities: visual and inertial. Visual-inertial navigation is a typical state estimation problem, that estimates ego-motion and local map given measurement data from the camera and IMU. Ego-motion can be quantitatively described as robot states normally involving translation, rotation, and velocity, evenly higher time derivatives (acceleration, etc.) in certain reference coordinates over time\cite{RN517}. In terms of local map representation, VIN usually quantifies the geometric information of local map in the form of 3D point clouds, meshes, voxels, or signed distance functions (SDF)\cite{RN631} with different capabilities and efficiencies. 

\begin{figure}[h]
    \centering
    \includegraphics[width=0.9\textwidth]{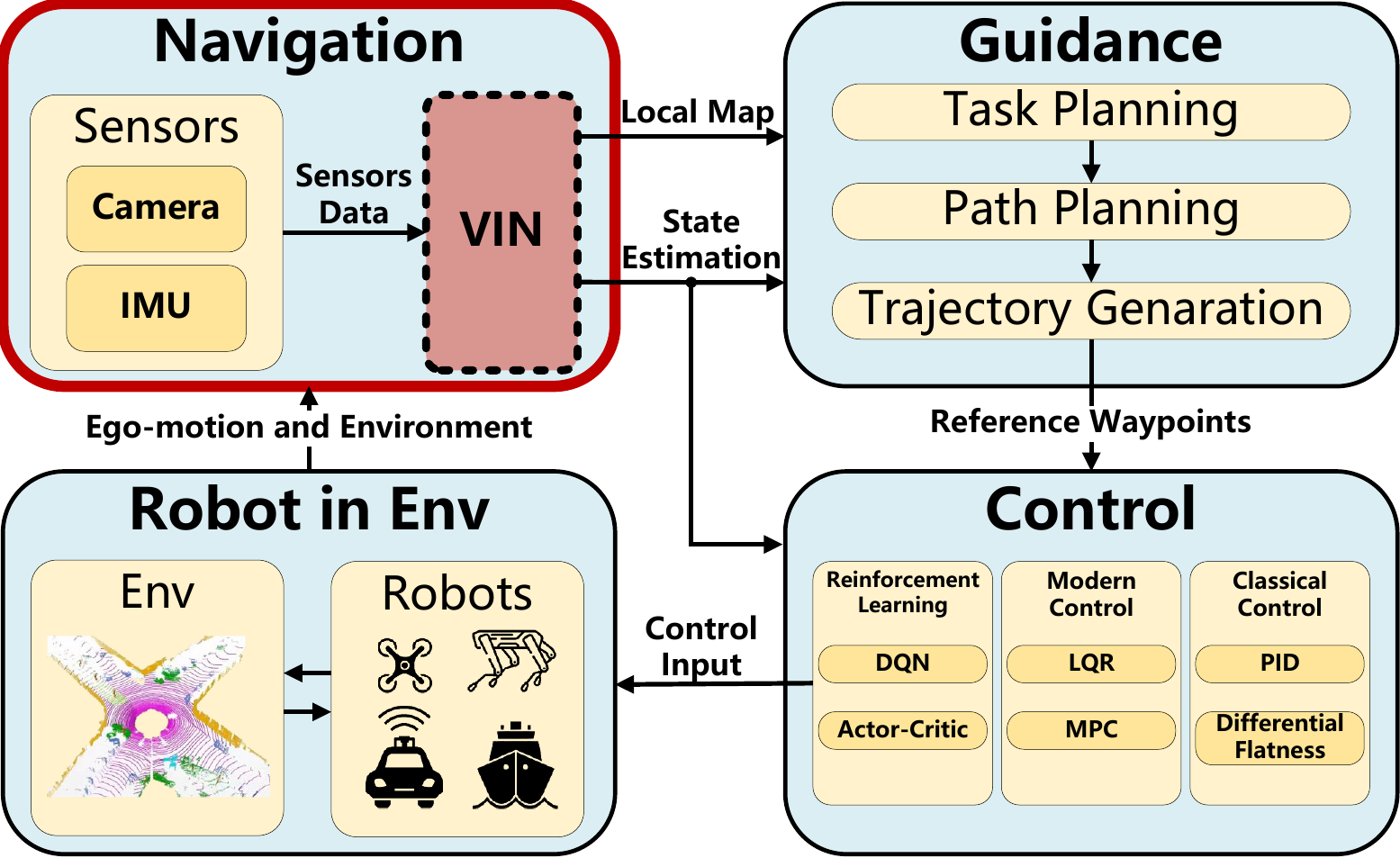}
    \caption{VIN in general guidance, navigation, and control (GNC) framework}
    \label{fig:1}
\end{figure}

% WHY SLAM
For robot navigation, VIN provides both robot's current state and local map, which are essential for path planning, obstacle avoidance, and real-time control. As a popular navigation technique, VIN greatly contributes to the general guidance, navigation, and control (GNC) systems shown in Figure \ref{fig:1}. In particular, since VIN uses onboard sensors, such sensory modalities are critically essential in the absence of external positioning systems like GPS, UWB, and visual motion capture systems. As such, VIN is widely used for indoor localization, underwater exploration, city reconstruction, and search-and-rescue. 

% Why VI

Compared to some GPS-free navigation system that uses only cameras and lidars, visual-inertial fusion improves accuracy, robustness, and efficiency of estimation. Visual-only systems like visual simultaneous localization and mapping(V-SLAM) require static scenes, sufficient illumination, Lambertian textured surface (no transparent or reflective), rich texture, and scene overlap. However, as a proprioceptive sensor, IMU is not limited by the above conditions by its internal mechanism that reads local linear acceleration and angular velocity. Micro-electro-mechanical-system (MEMS) IMU, in particular, has become ubiquitous in many robotics applications such as micro aerial vehicles (MAVs)\cite{RN94}\cite{RN767}, autonomous underwater vehicles (AUVs)\cite{RN760}, autonomous ground vehicles (AGVs)\cite{RN769}, quadruped robot\cite{RN1158}, spacecraft\cite{mourikis2009vision}, and mobile devices\cite{RN556} due to its low cost and lightweight. However, IMU uses the dead reckoning technique for estimation which leads to drift over time caused by integrating the bias and noisy measurements. Two types of sensory modalities compensate for each other, IMU won't affect by visual extreme conditions including motion blur, low texture, and illumination change, and provides high rate measurements, whereas cameras are more accurate at slow motion with no drift in static scenes.

% outline of this paper

\section{States and Models in VIN}
As a typical state estimation problem, VIN is to estimate the robot motion and local map given IMU kinematics model and camera measurement model. The robot's current ego-motion can be characterized as the pose at the current moment and its time derivatives (velocity and acceleration). The metric information of a local map can be represented as 3D points, lines, or surfaces distributed in 3D space, where semantic information acted as "category signatures" tagged on these metric representations. In this work, we focus on metric information and explore works that use 3D points as local map landmarks. Poses and 3D landmarks are normally of interest to VIN state estimation. In VIN, some supporting parameters are also needed for estimating poses and 3D landmarks, including some time-varying quantities like velocity, IMU bias, gravity direction in the local frame, and some time-invariant quantities like camera intrinsic, and camera-IMU extrinsic, etc. A typical state estimation model is shown in Equation \ref{eq:0a} and \ref{eq:0b} as dynamic and measurement models respectively. In many VIN works\cite{RN397}\cite{RN91}\cite{RN820}, systems are modeled as hybrid systems\cite{RN771} with continuous-time dynamic and discrete-time measurements. 
\begin{subequations}
\begin{align}
&\text{\textbf{Continuous-time Dynamic Model:}}&& \dot{\mathbf{x}}=\mathbf{f}(\mathbf{x},\tilde{\mathbf{u}},\mathbf{w})              \label{eq:0a} \\
&\text{\textbf{Discrete-time Measurement Model:}}&& \tilde{\mathbf{z}}_k=\mathbf{h}(\mathbf{x}_k)+\mathbf{v}\label{eq:0b}
\end{align}
\end{subequations}

In VIN, poses, landmarks, time-varying quantities(velocity, IMU bias, etc.), and sometimes time-invariant quantities (online calibration) are contained in state vector $\mathbf{x}$, the dynamic function $\mathbf{f}(\cdot)$ is represented by IMU kinematics with linear acceleration and angular velocity measurement as control input $\tilde{\mathbf{u}}$. The measurement model $\mathbf{h}(\cdot)$ is normally a camera measurement model with 2D image features as output $\tilde{\mathbf{z}}_k$. $\mathbf{w}$ and $\mathbf{v}$ are process noise (noises from IMU measurements) and measurement noise (noises from camera measurements) respectively. In this section, we first define states and reference frames in VIN problem and discuss the parametrization of quantities involved in VIN. Lastly, we briefly derive IMU kinematics model and camera measurement model and relate them as factors in the factor graph for estimation methods clarification in Section \ref{abcc}.

\subsection{Reference Frames and States in VIN}
\begin{figure}[h]
    \centering
    \includegraphics[width=0.5\textwidth]{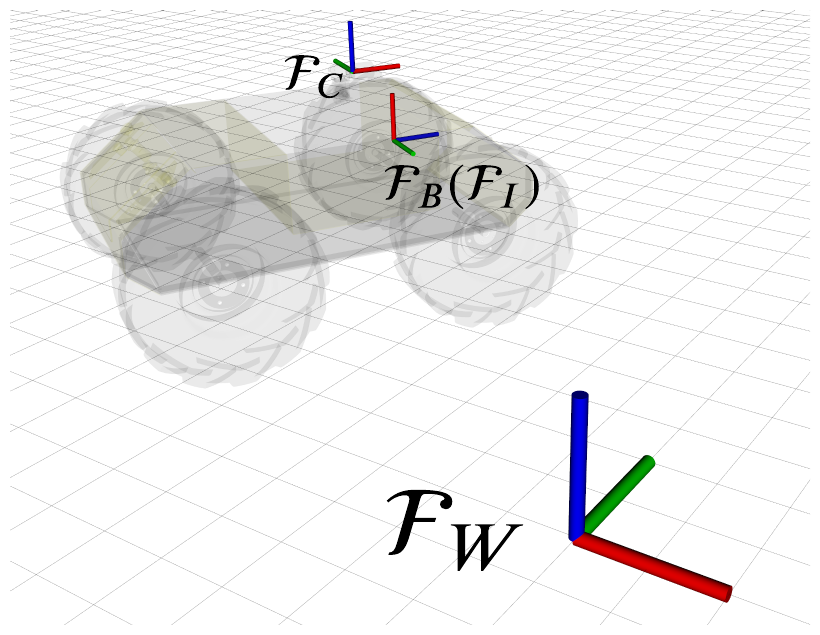}
    \caption{Fixed world frame and moving body frame with rigidly mounted sensor frames}
    \label{fig:3}
\end{figure}

Geometric quantities such as poses and landmarks are relative quantities that depend on the frame of reference (coordinate). It is critical to firmly identify different frames. These include a \textit{fixed} world frame $\bm{\mathcal{F}_{W}}$ (global frame, map frame, or inertial frame), a \textit{moving} body frame $\bm{\mathcal{F}_{B}}$ (robot frame or vehicle frame), some sensors frames (rigidly installed on the robot) like camera frames $\bm{\mathcal{F}_{C}}$ and IMU frames $\bm{\mathcal{F}_{I}}$. In many VIN cases, the IMU frame is coincident with the body frame as shown in Figure \ref{fig:3}, where $X$, $Y$, and $Z$ axis are in red, green, and blue respectively.

Normally, the final result of VIN including a pose track and a local map is expressed in an arbitrary stationary world frame. The pose of a robot at each time-step is expressed in translation and rotation $({}_{W}\bm{t}_{WB},\bm{R}_{WB})$ concerning the transformation between body frame and world frame. In world-centric framework, The 3D landmarks are also expressed in world frame. In many VIN practices, the world frame is set to the first camera frame's corresponding IMU frame $\bm{\mathcal{F}_{I_0}}$. This allows zero initial pose uncertainty, which reduces the level of uncertainty and increases the consistency of the estimate\cite{RN1206}\cite{RN962}.

 \begin{figure}[h]
     \centering
     \includegraphics[width=0.8\textwidth]{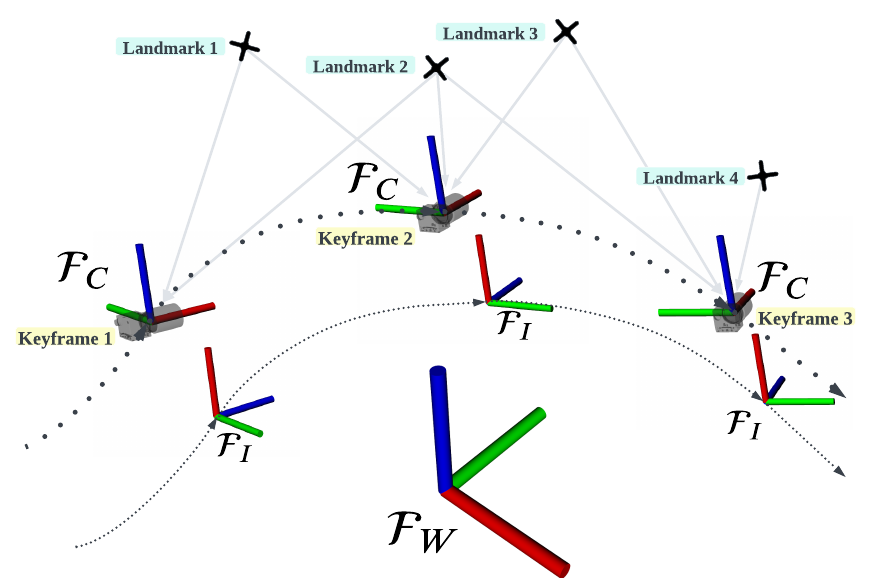}
     \caption{Landmarks and Camera-IMU poses along discrete measurement trajectory}
     \label{fig:6}
 \end{figure}

\begin{figure}[h]
     \centering
     \begin{subfigure}[b]{0.22\textwidth}
         \centering
         \includegraphics[width=\textwidth]{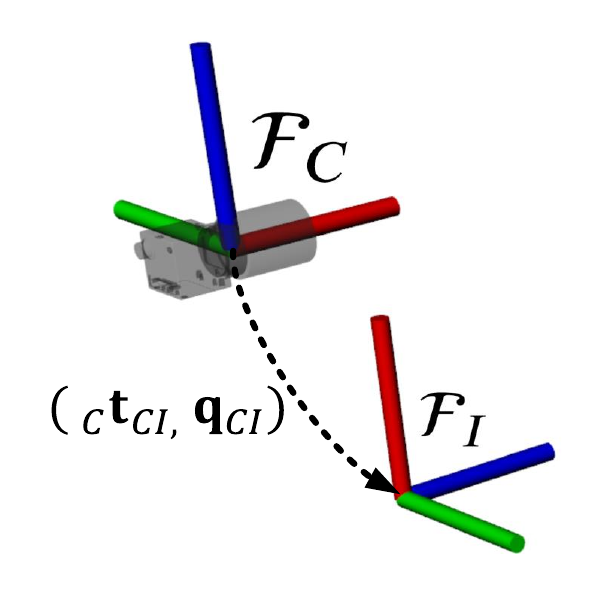}
         \caption{Camera-IMU Extrinsic}
         \label{ic:in}
     \end{subfigure}
     % \hfill
     \begin{subfigure}[b]{0.77\textwidth}
         \centering
         \includegraphics[width=\textwidth]{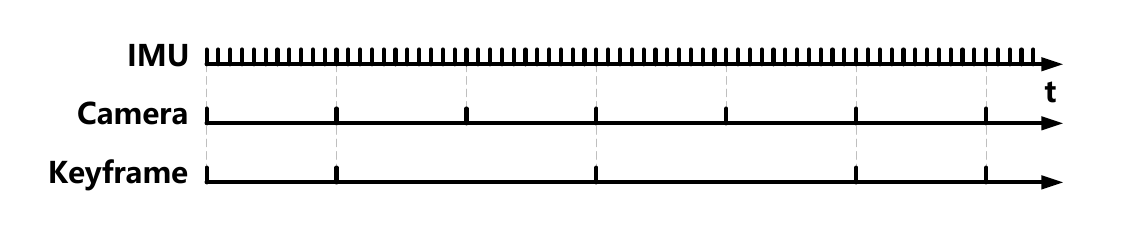}
         \caption{Time-synchronized IMU and camera measurement rate with possible keyframes selection}
         \label{fig:66}
     \end{subfigure}
        \caption{Spatial and Temporal difference between Camera and IMU}
        \label{ic:66}
\end{figure}

The visual measurements of a moving camera are shown in a sequence of image frames that are expressed in a 2D image coordinate. This 2D image coordinate is parallel to the y-z plane of the 3D camera frame $\bm{\mathcal{F}_{C}}$, which is rigidly mounted on the robot, as does the IMU frame $\bm{\mathcal{F}_{I}}$. Figure \ref{fig:6} shows sparse landmark features,   and discrete visual-inertial measurements along trajectories, whose sampling rate is indicated by the density of dashed lines. IMU and camera, as a multi-sensor system, exist temporal and spatial differences. The temporal difference is mainly reflected in their different measurement frequency and latency, and the spatial difference is caused by the fixed installation displacement of these two sensors (camera-IMU extrinsic as shown in Figure \ref{ic:in}). The measurement sampling rate of the IMU (100-1000Hz) is much faster than that of the camera (10-80Hz). Due to the difference in sensors' latency, there exists temporal misalignment (time offset) between IMU and camera measurements. This temporal misalignment can be resolved by temporal calibration\cite{RN747}\cite{RN407}\cite{RN92}\cite{RN795} that estimates the time offset or hardware synchronization\cite{RN419}, Figure \ref{fig:66} shows the periodic time-synchronized IMU and camera measurement timestamps with possible keyframe selection. Camera-IMU extrinsic can be identified by offline and online spatial calibration, which we discuss later in this paper. For now, we assume IMU and camera data are time-synchronized and camera-IMU extrinsic is known. 

 % \begin{figure}[h]
 %     \centering
 %     \includegraphics[width=0.9\textwidth]{time.pdf}
 %     \caption{Time-synchronized IMU and camera measurement rate with possible keyframes selection}
 %     \label{fig:66}
 % \end{figure}

 % Note that the IMU has a higher sampling rate than the camera. Therefore, in filter-based estimation, the estimated robot pose between two consecutive camera measurements can only be propagated through the IMU dynamic model and then updated when a new camera measurement is obtained.
The states of interest in VIN normally consist of poses and landmarks. In structureless visual-inertial estimation, landmarks are excluded from the state of interest. For filter-based estimation, the estimated state usually contains only one pose at the current moment, while in an optimization-based framework, a fixed or incremental number of poses will be involved in the state. In this work, we adopt Carlone $et$ $al.$\cite{RN1073} convention, which classifies \textit{target} variables and \textit{support} variables in the state variables, where target variables are normally of interest variables and support variables are essential for estimation of target variables. In VIN, target variables are the poses (translation and rotation) of a robot. Support variables contain the velocity and bias of IMU. Landmarks can be target or support variables depending on the focus of estimation. For example, in visual-inertial odometry (VIO), the current state of the robot including pose and its time derivatives are the most concerned, whereas, in visual-inertial simultaneous localization and mapping (VI-SLAM), accurate estimation of both landmarks and poses throughout the trajectory is required.

Note that the IMU dynamic model and the camera model are based on IMU and camera measurement respectively, in which case the maximum state estimation rate of VIN is bounded by the sensor measurement rate of the camera and IMU. A new state estimate can be computed whenever a new IMU or camera measurement arrives. However, the frequency of IMU measurements is high, which will result in a high computational burden and the prediction of estimation is prone to drift before a new camera measurement arrives. In this case, the poses' estimate of the robot is normally considered with the corresponding camera frames instead of all IMU frames, which means the poses in states are usually poses of camera frames' corresponding IMU frames (assume camera-imu extrinsic is known). 

For generality, we define active states in time $k$ in VIN as,
% \begin{subequations}
% \begin{align}
% &\text{\textbf{States in time $k$:}}&& \mathcal{X}_k=\left\{\mathcal{F}_k, \mathcal{L}_k\right\}               \label{eq:1aa} 
% \end{align}
% \end{subequations}
\[\mathcal{X}_k=\left\{\mathcal{I}_k, \mathcal{L}_k\right\} \]

where states in time $k$ consist of a set of IMU frames $\mathcal{I}_k$ and a set of 3D landmarks $\mathcal{L}_k$.
We assume all camera frames have corresponding time-synchronized IMU frames. The selections and number of frames are different depending on estimation frameworks, we here show some of the typical cases of frames' composition.
\begin{itemize}
    \item Filter-based framework: 
    
    Only contain one most recent camera frame, or the latest selected keyframe, or the last IMU frame in time $k$.
    \[\mathcal{I}_k=\left\{\mathbf{F}_k\right\}\text{, or} \left\{\mathbf{K}_k\right\}\text{, or}\left\{\mathbf{I}_k\right\}\]
    where $\mathbf{F}_k$ denotes the last camera frame corresponded IMU frame in time $k$, $\mathbf{K}_k$ denotes the last camera keyframe corresponded IMU frame in time $k$, $\mathbf{I}_k$ denotes the last IMU frame in time $k$.
    \item Fixed-Lag Smoother framework: 
    
    Normally contain a fixed sliding window of recent frames and consecutive keyframes in time $k$.
    \[\mathcal{I}_k=\left\{\mathbf{K}_{k-m},\cdots,\mathbf{K}_k,\mathbf{F}_{k-n},\cdots,\mathbf{F}_k\right\}\]
    where $\left\{\mathbf{K}_{k-m},\cdots,\mathbf{K}_k\right\}$ denotes $m+1$ recent consecutive keyframes corresponded IMU frames in time $k$, $\left\{\mathbf{F}_{k-n},\cdots,\mathbf{F}_k\right\}$ denotes $n+1$ recent camera frames corresponded IMU frames in time $k$. There is no overlap between keyframes and recent frames, a new camera frame will be selected as a new keyframe if it passes the keyframe selection test, otherwise, it will be a new recent frame. 
    \item Full batch optimization-based framework: 
    
    Contain the entire keyframes in time $k$ that incremented over time.
    \[\mathcal{I}_k=\left\{\mathbf{K}_{1},\cdots,\mathbf{K}_k\right\}\]
    \item Covisibility-based framework: 
    
    Contain frames that share sufficient common features with the most recent keyframe in time $k$ (covisibility graph). This may involve past frames selected independently of time.
     \[\mathcal{I}_k=\left\{\mathbf{K}_{k},\mathbf{K}^k_{c_1},\cdots,\mathbf{K}^k_{c_o}\right\}\]
     where $\mathbf{K}^k_{c_1},\cdots,\mathbf{K}^k_{c_o}$ denote a set of camera frames corresponded IMU frames in the covisibility graph of the last keyframe $\mathbf{K}_{k}$.
\end{itemize}

Note that the \textit{keyframes} mentioned above are carefully selected from camera frames that exhibit good feature tracking quality and significant parallax with previous keyframes. The purpose of selecting keyframes from camera frames is to reduce the uncertainty of initialization and to improve the efficiency of graph-based optimization by preserving the sparsity that only a subset of camera frames is maintained. 

Recent IMU frames $\mathbf{I}$, camera frames' corresponding IMU frames $\mathbf{F}$, and camera keyframes' corresponding IMU frames $\mathbf{K}$ contain pose information of the IMU frame and its direct support variables. In VIN, these IMU frames can be represented in IMU state $\mathbf{x}_I$, where camera-IMU extrinsic $\left({ }_{C} \mathbf{t}_{CI}, \mathbf{q}_{CI}\right)$ is assumed given and camera-IMU measurements are time-synchronized. For visual-inertial state estimation with a monocular camera, an IMU state is generally defined as followed,

\[\mathbf{x}_I=\left[{ }_{W} \mathbf{t}_{WI}^\intercal, \mathbf{q}_{W I}^\intercal,{ }_{W} \mathbf{v}_{W I}^\intercal, \mathbf{b}_{g}^\intercal, \mathbf{b}_{a}^\intercal\right]^\intercal\]
where
\begin{conditions*}
{ }_{W} \mathbf{t}_{WI}\in\mathbb{R}^{3}& Position of IMU frame origin wrt. world frame expressed in world frame\\
\mathbf{q}_{W I}\in \mathbb{S}^{3}& Rotation from IMU frame to world frame in Hamiltonian unit quaternion\\
{ }_{W} \mathbf{v}_{W I}\in\mathbb{R}^{3}& Velocity of IMU frame wrt. world frame expressed in world frame\\
\mathbf{b}_{g}\in\mathbb{R}^{3} &Gyro bias of IMU frame wrt. world frame expressed in IMU frame\\
\mathbf{b}_{a}\in\mathbb{R}^{3} &Accelerometer bias of IMU frame wrt. world frame expressed in IMU frame
\end{conditions*}

The set of 3D landmarks $\mathcal{L}_k$ in time $k$ contains all of the landmarks observed in the camera frames corresponding to the IMU frames $\mathcal{I}_k$. 
%为了清晰的区别在不同frame的feature, 定义一帧内观测的2d特征点，可能没有对应的3D点 
For generality, we define
\[\mathcal{L}_k=\left\{ \mathbf{L}^k_1,\cdots,\mathbf{L}^k_{l_k}\right\}\]
where $\mathbf{L}^k_j\;j=1,\cdots, l_k$ denotes a single 3D landmark, $l_k$ denotes the number of active 3D landmarks in $\mathcal{I}_k$. Each 3D landmark normally contains the position and indexing information. Position information can be straightforwardly presented in 3D Euclidean space, more parameterizations are presented in Section \ref{s32}. The indexing information for a 3D landmark contains a "serial number" that corresponds to the 2D feature in the set of camera frames that observed the landmark, and the indices of this set of camera frames. This "serial number" is a feature descriptor like ORB\cite{RN541} or BRIEF\cite{RN543}, etc. in descriptor-based data association. 

\subsection{Related Quantities in VIN and thier Parametrizations}\label{s32}

In this section, we discuss all relevant quantities involved in VIN estimation and the properties of their typical representations (parametrization). Pose and landmark is the most concerned quantities in VIN. The pose of frames consists of translation and rotation. We adopt Furgale's conversion\cite{RN1217}, where coordinates of 3-vectors in VIN like translation, velocity, and acceleration need three elements of decoration to clearly specify, which contain a physical quantity from one frame to another in addition to being represented in a specific reference frame. For rotation, two elements are required to represent the origin and destination frame. Assume for the moment that poses and landmarks are all expressed in the world frame. 

Parametrizations define a set of representations of the same quantity, like a 2D coordinate can be represented in both Polar and Cartesian Coordinates. However, some limitations come with certain parametrization, like singularities and constraints. The translation lives in vector space with no singularity and constraints. However, the rotation has several forms, including rotation matrix, Euler angle, axis-angle, rotation vector, and unit quaternion shown in the below notation tree. The summary of their properties is shown in Table \ref{tab:my_label}. Unit quaternion compared to other parametrizations, contains fewer constraints and no singularity becoming a "standard" way to represent rotation. Since rotation is not in vector space but in a smooth manifold (Lie group), conventional additive and subtractive operations in vector space cannot be used. In this case, in order to use canonical filtering and optimization methods built on vector space operations, it is necessary to transform the quantities in Lie groups into operable vector spaces (Lie algebras). We will discuss this later in Section \ref{onm}. 
\begin{multicols}{2}
\noindent
    \begin{forest}
  for tree={
    font=\ttfamily,
    inner sep = 1pt,
    grow'=0,
    child anchor=west,
    parent anchor=south,
    anchor=west,
    calign=first,
    edge path={
      \noexpand\path [draw, \forestoption{edge}]
      (!u.south west) +(7.5pt,0) |- node[fill,inner sep=1.25pt] {} (.child anchor)\forestoption{edge label};
    },
    before typesetting nodes={
      if n=1
        {insert before={[,phantom]}}
        {}
    },
    fit=band,
    before computing xy={l=15pt},
  }
[Pose
  [Translation
    [3D Euclidean Vector ${ }_{W} \mathbf{t}_{WI}\in\mathbb{R}^{3}$]
  ]
  [Rotation
    [Rotation Matrix $\mathbf{R}_{WI}\in\mathrm{SO}(3)$]
    [{Euler Angle $(p,q,r)_{WI}\in\mathbb{R}^{3}$}]
    [{Axis-angle $(\mathbf{n},\theta)_{WI}\in\mathbb{S}^{2}\times\mathbb{R}$}]
    [Rotation Vector $\boldsymbol{\theta}_{WI}\in\mathbb{R}^{3}$]
    [Unit Quaternion $\mathbf{q}_{WI}\in\mathbb{S}^{3}$]
  ]
]
\end{forest}
\columnbreak

\noindent
\begin{forest}
  for tree={
    font=\ttfamily,
    inner sep = 1pt,
    grow'=0,
    child anchor=west,
    parent anchor=south,
    anchor=west,
    calign=first,
    l sep=0pt,
    edge path={
      \noexpand\path [draw, \forestoption{edge}]
      (!u.south west) +(5pt,0) |- node[fill,inner sep=1.25pt] {} (.child anchor)\forestoption{edge label};
    },
    before typesetting nodes={
      if n=1
        {insert before={[,phantom]}}
        {}
    },
    fit=band,
    before computing xy={l=15pt},
    % ！！！！！
    if level>=2{draw=none,
           text width=5em,
           align=left}{},
  }
[Landmark
  [3D Landmark
    [{Euclidean Point $(x,y,z)\in\mathbb{R}^{3}$}]
    [{Homogeneous Point $(m_x,m_y,m_z,\rho)\in\mathbb{P}^{3}$}]
    [{Inverse Depth Parametrization\\$(x_0,y_0,z_0,\varepsilon,\alpha,\rho)\in\mathbb{R}^{6}$}]
  ]
  [2D Feature
    [{Pixel Location $(u,v)\in\mathbb{R}^{2}$}]
    [{Elevation and Azimuth Angle $(\varepsilon,\alpha)\in\mathbb{R}^{2}$}]
    [{3D Unit Vector $(n_x,n_y,n_z)\in\mathbb{S}^{2}$}]
  ]
]
\end{forest}
\end{multicols}
\begin{table}[h]
    \centering
    \begin{tabular}{c|c|c|c|c}
    &Dimension&Constraints&Direct Composition&Singularity  \\
    \hline
         Rotation matrix& $9$&$6$&Matrix Multiplication&None\\
         Euler angle& $3$&$0$&None&Gimbal lock\\
         Axis-angle & $4$&$1$&Compositable over Single Axis& $\theta=0$ and $\pi$\\
         Rotation Vector & $3$&$0$&Compositable over Single Axis& None\\
         Unit Quaternion & $4$&$1$&Quaternion Multiplication& None\\
    \end{tabular}
    \caption{Rotation Parametrizations}
    \label{tab:my_label}
\end{table}

Point landmarks in VIN exist in 3D space and are observed in their 2D projection. Geometrically, the camera is a bearing sensor, and distances to landmarks cannot be measured with a single frame. 3D landmarks in Euclidean space cannot be initialized at the first observation. In this case, the concept of undelayed landmark initialization (ULI) is introduced by Solà\cite{RN1213}, which represents the unmeasured DOF by a Gaussian prior. This involves two important  ULI landmark parametrizations: Homogeneous points (HP) and inverse depth parametrization (IDP). HP lives in 4-vector projective space, where $(m_x,m_y,m_z)$ denotes the 3D vector toward the landmark and the scalar $\rho$ is proportional to inverse distance. IDP is introduced by Montiel $et$ $al.$ in \cite{RN686} and extended by Civera in \cite{RN1214}. IDP maintains relatively low non-linearity in measurement equation at low and high feature parallax compared to Euclidean parametrization. Compared to HP, IDP adds a predicted first-seen frame as an anchor, which allows to represent landmark uncertainty locally, thus reducing the accumulated linearization error\cite{RN7761}. In \cite{RN1223}, Solà $et$ $al.$ show that IDP establishes better consistency in monocular EKF-SLAM compared to HP. 2D features are originally measured in the image coordinate with the pixel unit. However, in pixel coordinate parameterization, the same pixel coordinates represent different bearing directions in different cameras (perspective, refraction, or catadioptric cameras). For generality, the 3D bearing unit vector is used for representing the 2D features in the camera frame, it has no singular configurations compared to elevation and azimuth angle (north and south poles).

Other time-varying quantities like velocity and acceleration support the estimation of poses, whereas IMU bias, gravity, and earth rotation effect affect the estimate of velocity and acceleration. In addition, in the case of the monocular camera, scale ambiguity also needs to be considered. Local gravity and earth angular velocity expressed in IMU frame are fixed-length vectors ${ }_{I}\mathbf{g}=g{}_{I} \hat{\mathbf{g}}$ and ${}_{I}\boldsymbol{\omega}_E=\omega_\epsilon{ }_{I} \hat{\boldsymbol{\omega}}_E$, where $g\approx9.81 \mathrm{m/s^2}$ and $\omega_\epsilon\approx7.29\times10^{-5} \mathrm{rad/s}$ with 3D unit direction vectors ${ }_{I} \hat{\mathbf{g}}$ and ${ }_{I} \hat{\boldsymbol{\omega}}_E$, symbol $\hat{\{\cdot\}}$ indicates unit vector. All other quantities live in 3D vector space. Note angular velocity and linear acceleration are measured by gyroscope and accelerometer respectively. Earth rotation effect is measurable when using high-end IMUs, but is negligible in many VINs using low-cost IMUs. For simplicity, we neglect it in this tutorial, for considering this effect, we refer readers to \cite{RN397}\cite{RN517}.
%如果按顺序估计，应该首先是 Bias+gravity -> velocity and acceleration -> pose
\begin{multicols}{2}
\noindent
\begin{forest}
  for tree={
    font=\ttfamily,
    inner sep = 1pt,
    grow'=0,
    child anchor=west,
    parent anchor=south,
    anchor=west,
    calign=first,
    edge path={
      \noexpand\path [draw, \forestoption{edge}]
      (!u.south west) +(7.5pt,0) |- node[fill,inner sep=1.25pt] {} (.child anchor)\forestoption{edge label};
    },
    before typesetting nodes={
      if n=1
        {insert before={[,phantom]}}
        {}
    },
    fit=band,
    before computing xy={l=15pt},
  }
[Velocity and Acceleration
  [Linear Velocity ${ }_{W} \mathbf{v}_{W I}\in \mathbb{R}^{3}$
  ]
  [Angular Velocity ${}_{I}\boldsymbol{\omega}_{WI}\in\mathbb{R}^{3}$
  ]
  [Linear Acceleration ${}_{I}\boldsymbol{a}_{WI}\in\mathbb{R}^{3}$
  ]
]
\end{forest}
\columnbreak

\noindent
\begin{forest}
  for tree={
    font=\ttfamily,
    inner sep = 1pt,
    grow'=0,
    child anchor=west,
    parent anchor=south,
    anchor=west,
    calign=first,
    edge path={
      \noexpand\path [draw, \forestoption{edge}]
      (!u.south west) +(7.5pt,0) |- node[fill,inner sep=1.25pt] {} (.child anchor)\forestoption{edge label};
    },
    before typesetting nodes={
      if n=1
        {insert before={[,phantom]}}
        {}
    },
    fit=band,
    before computing xy={l=15pt},
  }
[{IMU Bias, Gravity, Scale and Earth Rate}
  [Gyroscope Bias $\mathbf{b}_{g}\in\mathbb{R}^{3}$
  ]
  [Accelerometer Bias $\mathbf{b}_{a}\in\mathbb{R}^{3}$
  ]
  [Local Gravity Direction ${ }_{I} \hat{\mathbf{g}}\in\mathbb{S}^{2}$
  ]
  [Monocular Scale $s\in\mathbb{R}$
  ]
  [Local Earth Rate Direction ${ }_{I} \hat{\boldsymbol{\omega}}_E\in\mathbb{S}^{2}$
  ]
]
\end{forest}
\end{multicols}

Besides time-varying quantities that must be estimated online, there are time-invariant quantities that can be estimated offline or online (out-of-box operation). 
These quantities are associated with the sensors properties of the camera and IMU, which can be classified into the camera, IMU, and camera-IMU related quantities shown in the below notation tree. For camera-related quantities, camera intrinsic should be considered since images are rendered in image coordinates in pixel units with lens distortion. Some VIN systems\cite{RN445}\cite{RN76}\cite{RN77} rely on the photometric intensity of feature pixels (direct methods), in which case time-varying auto exposure time, camera response function and the attenuation factors due to vignetting are also needed to be calibrated\cite{RN301}\cite{RN1227} for better direct feature tracking. For IMU-related quantities, the effects of axis misalignment, scale factor errors, and linear acceleration on the gyroscope should be considered, especially for low-cost, consumer-grade MEMS IMUs. For camera-IMU-related quantities, the temporal and spatial differences between these two sensors should be considered including camera-IMU extrinsic and their time offset. In a multi-camera multi-IMU setup, these differences in camera-camera and IMU-IMU should also be calibrated.

% \hspace{1cm}

\noindent
\begin{forest}
  for tree={
    font=\ttfamily,
    inner sep = 1pt,
    grow'=0,
    child anchor=west,
    parent anchor=south,
    anchor=west,
    calign=first,
    edge path={
      \noexpand\path [draw, \forestoption{edge}]
      (!u.south west) +(7.5pt,0) |- node[fill,inner sep=1.25pt] {} (.child anchor)\forestoption{edge label};
    },
    before typesetting nodes={
      if n=1
        {insert before={[,phantom]}}
        {}
    },
    fit=band,
    before computing xy={l=15pt},
    % ！！！！！
    if level>=1{draw=none,
           text width=5em,
           align=left}{},
  }
[Relevant Time Invariant Quantities
  [Camera Intrinsic
  [{Principal Point $(c_x,c_y)\in\mathbb{R}^2$}]
  [{Focal length $(f_x,f_y)\in\mathbb{R}^2$}]
  [{Radial and Tangential Distortion Parameters $(\mathbf{k},\mathbf{p})\in\mathbb{R}^6\times\mathbb{R}^2$}]
  ]
  [{IMU Axis misalignment, Scale factor, and Linear acceleration effect on gyro}
  [{Gyroscope and Accelerometer Axis misalignment $\mathbf{M}_g,\mathbf{M}_a\in\boldsymbol{L}^{3\times3}$}]
  [{Gyroscope and Accelerometer Scale factor $\mathbf{S}_g,\mathbf{S}_a\in\boldsymbol{\Lambda}^{3\times3}$}]
  [Linear acceleration effect on gyro $\mathbf{B}_g\in\mathbb{R}^{3\times3}$]
  ]
  [Camera-IMU Extrinsic and Time Offset
  [Camera-IMU Displacement ${ }_{C} \mathbf{t}_{CI}\in\mathbb{R}^{3}$]
  [Camera-IMU Orientation $\mathbf{q}_{CI}\in\mathbb{S}^{3}$]
  [Camera-IMU Time Offset $t_d\in\mathbb{R}$]
  ]
]
\end{forest}

The active states in VIN including poses-related variables and landmarks are relative quantities expressed in a relative reference frame. Based on the reference frame of the active state, VIN can be divided into world-centric and robot-centric frameworks. Landmarks are measured locally, presenting them on a local frame limits the level of uncertainty and thus reduces the linearization errors and increases consistency in EKF framework\cite{RN1206}\cite{RN962}\cite{RN1077}.

\hspace{1cm}

\noindent
\begin{forest}
  for tree={
    font=\ttfamily,
    grow'=0,
    child anchor=west,
    parent anchor=south,
    anchor=west,
    calign=first,
    edge path={
      \noexpand\path [draw, \forestoption{edge}]
      (!u.south west) +(7.5pt,0) |- node[fill,inner sep=1.25pt] {} (.child anchor)\forestoption{edge label};
    },
    before typesetting nodes={
      if n=1
        {insert before={[,phantom]}}
        {}
    },
    fit=band,
    before computing xy={l=15pt},
  }
[Reference Frame of Active States
  [World-Centirc ${}_{W}\{\cdot\}$
  ]
  [Robot-Centirc ${}_{I}\{\cdot\}$
  ]
]
\end{forest}
% Note that unit quaternions $S^{3}$ are used for expressing rotation, which are unit sphere in four dimensions. Unit quaternions have many advantages over other rotations representations including orthonormal rotation matrices, Euler angles, axis-angle and rotation vector\cite{RN507}. In this paper, the right-handed Hamiltonian unit quaternion convention is adopted.

% \[\mathbf{q}_{W I}=\underbrace{\begin{bmatrix}
% q_w\\
% {\color{black}q_x}\\
% {\color{black}q_y}\\
% {\color{black}q_z}
% \end{bmatrix}}_{\text{Vector form}}\in S^{3},\triangleq\underbrace{q_w+{\color{black}q_xi+q_yj+q_zk}}_{\text{Hypercomplex number form}}\in\mathbb{H}\]
% \[i^2=j^2=k^2=ijk=-1,\quad \Vert\mathbf{q}_{W I}\Vert=\sqrt{q_w^2+{\color{black}q_x^2+q_y^2+q_z^2}}=1\]
% where,
% \begin{conditions*}
% q_w\in\mathbb{R} & Real part of unit quaternion\\
% {\color{black}q_x,q_y,q_z}\in\mathbb{R} &  Imaginary part of unit quaternion
% \end{conditions*} 

\subsection{IMU and its Kinematic Model}
IMU, as a proprioceptive sensor, measures angular velocity ${}_{I}\tilde{\boldsymbol{\omega}}_{WI}$ and linear acceleration ${}_{I}\tilde{\boldsymbol{a}}_{WI}$ by gyroscope and accelerometer respectively. Generally, IMUs are often grouped into five categories in terms of performance: consumer-grade, tactical-grade, intermediate-grade, aviation-grade, and marine-grade with decreasing bias and random walk but dramatically increasing price\cite{RN12}. There are also four major types of IMUs based on their mechanism: mechanical gyroscopes, RLG (ring laser gyroscopes), FOG (fiber optic gyroscopes), and MEMS (micro-electro-mechanical systems)\cite{RN822}\cite{RN823}. Although MEMS IMUs normally perform in consumer-grade and tactical-grade, they are extremely low-cost, lightweight, and have low power consumption. 

A general IMU measurement model is shown in Equation \ref{eq:1a} and \ref{eq:1b}. However, consumer-grade IMUs often exist axis misalignment, scale factors, and linear acceleration effects on gyroscope. They can be estimated by IMU calibration\cite{RN408} and be used in a extended IMU measurement model\cite{RN1159}.

\begin{subequations}
\begin{align}
&\text{\textbf{Gyroscope Measurement:}}&& {}_{I}\tilde{\boldsymbol{\omega}}_{WI}={}_{I}\boldsymbol{\omega}_{WI}+\mathbf{b}_{g}+\mathbf{n}_{g}               \label{eq:1a} \\
&\text{\textbf{Accelerometer Measurement:}}&& {}_{I}\tilde{\boldsymbol{a}}_{WI}=\mathbf{R}_{I W}\; { }_{W} \boldsymbol{a}_{WI}+{ }_{I} \mathbf{g}+\mathbf{b}_{a}+\mathbf{n}_{a}  \label{eq:1b}
\end{align}
\end{subequations}
where
\begin{conditions*}
{}_{I}\tilde{\boldsymbol{\omega}}_{WI}\in\mathbb{R}^{3} & Measured angular velocity of IMU frame wrt. world frame expressed in IMU frame\\
{}_{I}\tilde{\boldsymbol{a}}_{WI}\in\mathbb{R}^{3} & Measured acceleration of IMU frame wrt. world frame expressed in IMU frame\\
{}_{I}\boldsymbol{\omega}_{WI}\in\mathbb{R}^{3} & True angular velocity of IMU frame wrt. world frame expressed in IMU frame\\
{ }_{W} \boldsymbol{a}_{WI}\in\mathbb{R}^{3} & True acceleration of IMU frame wrt. world frame expressed in IMU frame\\
\mathbf{n}_{g}\in\mathbb{R}^{3} & Gyroscope measurement noise (angular random walk)\\
\mathbf{n}_{a}\in\mathbb{R}^{3} & Accelerometer measurement noise (velocity random walk)\\
\mathbf{R}_{I W}\in \mathrm{SO}(3)& Rotation matrix from world frame to IMU frame\\
{ }_{I} \mathbf{g}\in g\cdot\mathbb{S}^{2}& Gravitational acceleration in IMU frame
\end{conditions*}
Define Hamiltonian unit quaternion to rotation matrix transformation,
\[\mathbf{q}=\begin{bmatrix}
    q_w\\
    q_x\\
    q_y\\
    q_z
\end{bmatrix}\in\mathbb{S}^3,\quad
\mathbf{R}=
\begin{bmatrix}
q_w^2+q_x^2-q_y^2-q_z^2 & 2\left(q_x q_y-q_w q_z\right) & 2\left(q_x q_z+q_w q_y\right) \\
2\left(q_x q_y+q_w q_z\right) & q_w^2-q_x^2+q_y^2-q_z^2 & 2\left(q_y q_z-q_w q_x\right) \\
2\left(q_x q_z-q_w q_y\right) & 2\left(q_y q_z+q_w q_x\right) & q_w^2-q_x^2-q_y^2+q_z^2
\end{bmatrix}\in\mathrm{SO}(3)\]

In VIN's IMU dynamic models, measurements of angular velocity ${}_{I}\tilde{\boldsymbol{\omega}}_{WI}$ and acceleration ${}_{I}\tilde{\boldsymbol{a}}_{WI}$ are normally treated as input. The IMU biases are modeled as {random walk processes}, driven by the white Gaussian noise. The continuous-time IMU dynamic is expressed in the nonlinear state space equation as shown, 
\begin{subequations}
\begin{align}
    &\text{\textbf{Translation:}}&&{}_{W}\mathbf{\dot{t}}_{WI}={ }_{W} \mathbf{v}_{WI}\label{5a}\\
    &\text{\textbf{Velocity:}}&&{ }_{W} \mathbf{\dot{v}}_{WI}={ }_{W} \boldsymbol{a}_{WI}\label{5c}\\
    &\text{\textbf{Rotation:}}&& \mathbf{\dot{q}}_{WI}=\frac{1}{2}\mathbf{q}_{WI}\otimes\begin{bmatrix}
        0\\
        {}_{I}\boldsymbol{\omega}_{WI}
    \end{bmatrix}
    \label{5b}\\
    &\text{\textbf{Gyroscope bias:}}&&\mathbf{\dot{b}}_{g}=\mathbf{n}_{\mathbf{b}_{g}}\label{5d}\\
    &\text{\textbf{Accelerometer bias:}}&&\mathbf{\dot{b}}_{a}=\mathbf{n}_{\mathbf{b}_{a}}\label{5e}
\end{align}
\end{subequations}
where 
\begin{conditions*}
\mathbf{n}_{\mathbf{b}_{g}}\in\mathbb{R}^{3} & Gyroscope random walk (rate random walk)\\
\mathbf{n}_{\mathbf{b}_{a}}\in\mathbb{R}^{3} & Accelerometer random walk (acceleration random walk)
\end{conditions*}
Since the angular velocity is expressed locally in IMU frame, Equation \ref{5b} can be obtained by quaternion multiplication, where the local angular velocity is the second term of the product in pure quaternion representation. The quaternion multiplication is shown in Equation \ref{5f},
\begin{equation}
    \mathbf{\dot{q}}_{WI}=\frac{1}{2}\mathbf{q}_{WI}\otimes\begin{bmatrix}
        0\\
        {}_{I}\boldsymbol{\omega}_{WI}
    \end{bmatrix}=\frac{1}{2}\boldsymbol{\Omega}\left(\begin{bmatrix}
        0\\
        {}_{I}\boldsymbol{\omega}_{WI}
    \end{bmatrix}\right)\mathbf{q}_{WI}\label{5f}
\end{equation}
where,
\[\begin{bmatrix}
        0\\
        \boldsymbol{\omega}
    \end{bmatrix}=\begin{bmatrix}
    0\\
    \omega_1\\
    \omega_2\\
    \omega_3
\end{bmatrix}\in\mathbb{H}_p,\quad\quad
\boldsymbol{\Omega}(\begin{bmatrix}
        0\\
        \boldsymbol{\omega}
    \end{bmatrix})=\begin{bmatrix}
0 & -\omega_1 & -\omega_2 & -\omega_3 \\
\omega_1 & 0 & -\omega_3 & \omega_2 \\
\omega_2 & \omega_3 & 0 & -\omega_1 \\
\omega_3 & -\omega_2 & \omega_1 & 0
\end{bmatrix}\in\mathbb{R}^{4\times4}\]

Note that gyroscope and accelerometer random walks rarely appear in the datasheet, but can be obtained by using Allan standard deviation\cite{RN748}\cite{RN745}\cite{RN827}. In this work, these four noises $\mathbf{n}_{g},\mathbf{n}_{a},\mathbf{n}_{\mathbf{b}_{g}},\mathbf{n}_{\mathbf{b}_{a}}$ are all assumed to be zero mean and uncorrelated white Gaussian processes. These noise parameters (noise covariance matrix $\mathbf{Q}$) should be determined offline or adaptively by sensor calibration.
In addition, for high-end IMUs (navigation grade or aviation grade), the earth rotation effect is measurable, and the IMU dynamics model considering this effect can be found in \cite{RN397}\cite{RN517}. By substituting Equation \ref{eq:1a} and \ref{eq:1b} into Equation \ref{5b} and \ref{5c} respectively, the non-linear continuous-time state space Equation of IMU dynamic can be obtained in Equation \ref{6a}.  

\begin{equation}
    \dot{\mathbf{x}}_I=\mathbf{f}(\mathbf{x}_I,\tilde{\mathbf{u}},\mathbf{w}) \label{6a}
\end{equation}
where
\begin{conditions*}
\mathbf{x}_I\in\mathbb{R}^{3}\times\mathbb{S}^{3}\times\mathbb{R}^{3}\times\mathbb{R}^{3}\times\mathbb{R}^{3} &IMU states, $({ }_{W} \mathbf{t}_{WI}, \mathbf{q}_{W I},{ }_{W} \mathbf{v}_{W I}, \mathbf{b}_{g}, \mathbf{b}_{a})$\\
\tilde{\mathbf{u}}\in\mathbb{R}^{3}\times\mathbb{R}^{3} & IMU measurements, $({}_{I}\tilde{\boldsymbol{\omega}}_{WI},{}_{I}\tilde{\boldsymbol{a}}_{WI})$\\
\mathbf{w}\in\mathbb{R}^{3}\times\mathbb{R}^{3}\times\mathbb{R}^{3}\times\mathbb{R}^{3} & IMU noises, $(\mathbf{n}_{g},\mathbf{n}_{a},\mathbf{n}_{\mathbf{b}_{g}},\mathbf{n}_{\mathbf{b}_{a}})$
\end{conditions*}

In state estimation, we can propagate the state estimate and its uncertainty through a dynamic model. To propagate state estimates in actual implementations, discretization of nonlinear continuous-time dynamic models is required. The discretization can be obtained by exact or numerical integration during period $[t_{k},t_{k+1}]$. For simplifying the notations, we denote IMU states at time step $k$ (time $t_k$) as $\mathbf{x}_{I_{k}}=\left[{}_{W}\mathbf{t}_{k}^\intercal,\mathbf{q}_{WI_{k}}^\intercal, { }_{W} \mathbf{v}_{k}^\intercal, \mathbf{b}_{g_{k}}^\intercal,\mathbf{b}_{a_{k}}^\intercal\right]^\intercal$ and IMU measurements at time step $k$ as $\tilde{\mathbf{u}}_k=\left[ {}_{I_k}\tilde{\boldsymbol{\omega}}^\intercal,{}_{I_k}\tilde{\boldsymbol{a}}^\intercal\right]^\intercal$, and define the periodic interval between time steps $\Delta=t_{k+1}-t_{k}$. 

For generality, we adopt the constant linear velocity model from \cite{RN742} (see the second term in RHS of Equation \ref{7a}). In addition, we assume that the direction of the angular velocity does not change in the interval\cite{RN1231}, as shown in Equation \ref{7b}. For the time-varying rotation axis of angular velocity, we refer readers to \cite{RN1231} and \cite{RN414}. As such, the discretized model with exact integration can be shown in the Equations \ref{7a}-\ref{7e}
% and also assume constant angular velocity measurement, where $\mathbf{q}(\cdot)$ transfer rotation vector to unit quaternion.

\begin{subequations}
\begin{flalign}
    &\text{\textbf{Translation:}}&&{}_{W}\mathbf{t}_{k+1}={}_{W}\mathbf{t}_{k}+{ }_{W} \mathbf{v}_{k}\Delta-\frac{1}{2}{ }_{W} \mathbf{g}\Delta^2+\mathbf{R}_{WI_{k}}\int\limits_{t_k}^{t_{k+1}}\int\limits_{t_k}^{\tau}\mathbf{R}_{I_{k}I_{t}}\left({}_{I_t}\tilde{\boldsymbol{a}}-\mathbf{b}_{a_{t}}-\mathbf{n}_{a} \right)\mathrm{d}t\mathrm{d}\tau\label{7a}\\
    % &\text{\textbf{Rotation:}}&& \mathbf{q}_{WI_{k+1}}=\mathbf{q}_{WI_{k}}\otimes\mathbf{q}\big(\left({}_{I_k}\tilde{\boldsymbol{\omega}}-\mathbf{b}_{g_k}-\mathbf{n}_{g}\right)\Delta\big)
     &\text{\textbf{Velocity:}}&&{ }_{W} \mathbf{v}_{k+1}={ }_{W} \mathbf{v}_{k}-{ }_{W} \mathbf{g}\Delta+\mathbf{R}_{WI_{k}}\int_{t_k}^{t_{k+1}}\mathbf{R}_{I_{k}I_{t}}\left({}_{I_t}\tilde{\boldsymbol{a}}-\mathbf{b}_{a_{t}}-\mathbf{n}_{a} \right)\mathrm{d}t\label{7c}\\
    &\text{\textbf{Rotation:}}&& \mathbf{q}_{WI_{k+1}}=\mathbf{q}_{WI_{k}}\otimes\exp \left(\begin{bmatrix}
        0\\
        \frac{1}{2} \int_{t_k}^{t_{k+1}} \left({}_{I_t}\tilde{\boldsymbol{\omega}}-\mathbf{b}_{g_t}-\mathbf{n}_{g}\right) \mathrm{d} t
    \end{bmatrix}\right)
    \label{7b}\\
    &\text{\textbf{Gyro. bias:}}&&\mathbf{b}_{g_{k+1}}=\mathbf{b}_{g_{k}}+\int^{t_{k+1}}_{t_k}\mathbf{n}_{\mathbf{b}_{g}}\mathrm{d}t\label{7d}\\
    &\text{\textbf{Acc. bias:}}&&\mathbf{b}_{a_{k+1}}=\mathbf{b}_{a_{t}}+\int_{t_k}^{t_{k+1}}\mathbf{n}_{\mathbf{b}_{a}}\mathrm{d}t\label{7e}
\end{flalign}
\end{subequations}

where $\exp(\cdot)$ is the quaternion exponential that maps the angular velocity in pure quaternion (Lie Algebra) to quaternion (Lie Group). 
\[\begin{bmatrix}
        0\\
        \frac{\boldsymbol{\omega}}{2}
    \end{bmatrix}=\frac{1}{2}\begin{bmatrix}
    0\\
    \omega_1\\
    \omega_2\\
    \omega_3
\end{bmatrix}\in\mathbb{H}_p,\quad\quad
\exp\left(\begin{bmatrix}
        0\\
        \frac{\boldsymbol{\omega}}{2}
    \end{bmatrix}\right)=\begin{cases}{\left[\cos \left(\frac{\|\boldsymbol{\omega}\|}{2}\right), \frac{\boldsymbol{\omega}^{\intercal}}{\|\boldsymbol{\omega}\|} \sin \left(\frac{\|\boldsymbol{\omega}\|}{2}\right)\right]^{\intercal}} &\in\mathbb{S}^{3} \text {, otherwise} \\ {\left[1, \frac{1}{2}\boldsymbol{\omega}^{\intercal} \right]^{\intercal}} &\in\mathbb{H} \text{, if }\|\boldsymbol{\omega}\| \rightarrow 0\end{cases}\]

However, the exact discretized model contains integral terms that may or may not have the corresponding closed-form solution in analytical form. Also, discrete IMU measurements are sampled at periodic time steps, so certain assumptions (piecewise constant or piecewise linear) should be made during interval integration. In this case, closed-form or numerical integration with certain assumptions is used to eliminate the integral terms. Note that in Equation \ref{7d} and \ref{7e} the integration of noises does not shift mean but increases the uncertainty of bias. Isolating the integral terms, we obtain the "preintegrated" IMU measurements as follows in Equations \ref{8a}-\ref{8c}, 
\begin{subequations}
\begin{align}
    \boldsymbol{\alpha}_{I_{k}I_{k+1}} &= \int_{t_k}^{t_{k+1}}\int_{t_k}^{\tau}\mathbf{R}_{I_{k}I_{t}}\left({}_{I_t}\tilde{\boldsymbol{a}}-\mathbf{b}_{a_{t}}-\mathbf{n}_{a} \right)\mathrm{d}t\mathrm{d}\tau\label{8a}\\
    \boldsymbol{\beta}_{I_{k}I_{k+1}} &= \int_{t_k}^{t_{k+1}}\mathbf{R}_{I_{k}I_{t}}\left({}_{I_t}\tilde{\boldsymbol{a}}-\mathbf{b}_{a_{t}}-\mathbf{n}_{a} \right)\mathrm{d}t\label{8b}\\
    \boldsymbol{\gamma}_{I_{k}I_{k+1}}&=\exp \left(\begin{bmatrix}
        0\\
        \frac{1}{2} \int_{t_k}^{t_{k+1}} \left({}_{I_t}\tilde{\boldsymbol{\omega}}-\mathbf{b}_{g_t}-\mathbf{n}_{g}\right) \mathrm{d} t
    \end{bmatrix}\right)\label{8c}
\end{align}
\end{subequations}

These three integral terms $(\boldsymbol{\alpha}_{I_{k}I_{k+1}},\boldsymbol{\beta}_{I_{k}I_{k+1}},\boldsymbol{\gamma}_{I_{k}I_{k+1}})$ can be preintegrated with zero initial conditions (identity for rotation) since they only depend on IMU measurements and bias in interval $[t_k,t_{k+1}]$.
They can be obtained by closed-form or numerical integration with certain piece-wise assumptions. The numerical integration is normally presented by Runge-Kutta integration in different orders($1^{\text{th}}$ order Euler, $2^{\text{nd}}$ order Tustin or mid-point, or $4^{\text{th}}$ order RK4) with increasing precision and computational cost. The closed-form integration is derived by Eckenhoff $et$ $al.$ in \cite{RN820}. In terms of the piece-wise assumptions during integration, Forster $et$ $al.$ assume piecewise constant $({}_{I}{\boldsymbol{\omega}}_{WI},{}_{W}{\boldsymbol{a}}_{WI})$ in \cite{RN505}\cite{RN400}, where Eckenhoff $et$ $al.$\cite{RN758}\cite{RN820} consider two models that assumes piecewise constant $({}_{I}\tilde{\boldsymbol{\omega}}_{WI},{}_{I}\tilde{\boldsymbol{a}}_{WI})$ and $({}_{I}\tilde{\boldsymbol{\omega}}_{WI},{}_{I}{\boldsymbol{a}}_{WI})$. Since preintegration terms also depend on the IMU bias, a first-order Taylor expansion on the bias linearization point is usually used to avoid re-preintegration when the bias linearization point changes\cite{RN820}\cite{RN505}\cite{RN91}.

By applying the closed-form or numerical integration with proper piece-wise assumption, a noise-free discretized model without integral terms $\mathbf{f}_{d}(\cdot)$ can be obtained to propagate state estimate in the filter-based framework or to construct the residual terms in the optimization-based framework in Equation \ref{9a} and \ref{9b} respectively. $\ominus$ denotes a generic minus operation.
\begin{subequations}
\begin{align}
    &\text{\textbf{State Propagation:}}&&\mathbf{x}_{I_{k+1}}=\mathbf{f}_{d}\left(\mathbf{x}_{I_{k}},\tilde{\mathbf{u}}_k,0\right)\label{9a}\\
    &\text{\textbf{State Residuals:}}&&\mathbf{e}_k=\mathbf{x}_{I_{k+1}}\ominus\mathbf{f}_{d}\left(\mathbf{x}_{I_{k}},\tilde{\mathbf{u}}_k,0\right) \label{9b}
\end{align}
\end{subequations}

Since the IMU measurement rate is much faster than that of the camera, the state propagation will take place several times before a new camera measurement arises for state update. In this case, we drop the conventional notion of the posteriori estimate $\{\cdot\}^{+}$ after update  and the priori estimate $\{\cdot\}^{-}$ through propagation.

To propagate uncertainty through dynamics, we adopt concepts from Solà's paper\cite{RN396} on error state dynamics, in which case, the true state $\mathbf{x}_{I}^{\text{true}}$ can be decomposed into a nominal state $\mathbf{x}_{I}$ containing a large signal and an error state $\delta\mathbf{x}_{I}$ containing a small signal.

\[\mathbf{x}_{I}^{\text{true}}=\mathbf{x}_{I}\oplus\delta\mathbf{x}_{I}\]

where $\oplus$ denotes a generic plus operation. Hence, there exist two dynamics: nominal state and error state dynamic, where the nominal state dynamic can be used for propagating the state estimate (mean), and the error state dynamic can be used for propagating uncertainty (covariance). The nominal state is propagated by $\mathbf{f}_{d}(\cdot)$. In order to propagate the error state, we first define the error state in VIN as,

\[\delta\mathbf{x}_I=[{ }_{W} \delta\mathbf{t}_{WI}^\intercal, \delta\boldsymbol{\theta}_{W I}^\intercal,{ }_{W} \delta\mathbf{v}_{W I}^\intercal, \delta\mathbf{b}_{g}^\intercal, \delta\mathbf{b}_{a}^\intercal]^\intercal\in\mathbb{R}^{3}\times\mathbb{R}^{3}\times\mathbb{R}^{3}\times\mathbb{R}^{3}\times\mathbb{R}^{3}\]
where
\begin{conditions*}
\delta\boldsymbol{\theta}_{W I}\in\mathbb{R}^{3} & Perturbation of rotation, which is Lie Algebra of unit quaternion in Cartesian vector space
\end{conditions*}
By using quaternion exponential $\exp(\cdot)$ that maps Lie Algebra to Lie Group, the composition of nominal and error state of rotation can be obtained as follows, 
\[\mathbf{q}^{\text{true}}=\mathbf{q}\otimes\exp\left(\begin{bmatrix}
        0\\
        \frac{\delta\boldsymbol{\theta}}{2}
    \end{bmatrix}\right)\]
Error state consists of only small signals, it can be considered as a perturbation of the nominal state. Thus, the error state dynamic preserves linearity. We adopt the linearized error-state process model in \cite{RN396} but drop the gravity term for generality.

\begin{subequations}
\begin{align}
    &\text{\textbf{Translation:}}&&{}_{W}\delta\mathbf{\dot{t}}_{WI}={ }_{W} \delta\mathbf{v}_{WI}\label{91a}\\
    &\text{\textbf{Velocity:}}&&{ }_{W} \delta\mathbf{\dot{v}}_{WI}=-\mathbf{R}_{WI}
    \lfloor{}_{I}\tilde{\boldsymbol{a}}_{WI}-\mathbf{b}_{a}\rfloor_\times\delta\boldsymbol{\theta}_{WI}-\mathbf{R}_{WI}\delta\mathbf{b}_{a}-\mathbf{R}_{WI}\mathbf{n}_{\mathbf{b}_{a}}
    \label{91c}\\
    &\text{\textbf{Rotation:}}&& \delta\dot{\boldsymbol{\theta}}_{WI}=-\lfloor{}_{I}\tilde{\boldsymbol{\omega}}_{WI}-\mathbf{b}_{g}\rfloor_\times\delta\boldsymbol{\theta}_{WI}-\delta\mathbf{b}_{g}-\mathbf{n}_{\mathbf{b}_{g}}
    \label{91b}\\
    &\text{\textbf{Gyroscope bias:}}&&\delta\mathbf{\dot{b}}_{g}=\mathbf{n}_{\mathbf{b}_{g}}\label{91d}\\
    &\text{\textbf{Accelerometer bias:}}&&\delta\mathbf{\dot{b}}_{a}=\mathbf{n}_{\mathbf{b}_{a}}\label{91e}
\end{align}
\end{subequations}

where the skew operation $\lfloor\cdot\rfloor_\times$  is defined as,
\[\boldsymbol{\omega}=\begin{bmatrix}
    \omega_1\\
    \omega_2\\
    \omega_3
\end{bmatrix}\in\mathbb{R}^3,\quad\quad
\lfloor\boldsymbol{\omega} \rfloor_\times=\begin{bmatrix}
  0 & -\omega_3 & \omega_2 \\
\omega_3 & 0 & -\omega_1 \\
-\omega_2 & \omega_1 & 0  
\end{bmatrix}\in\mathbb{R}^{3\times3}\]

A continuous-time linearization error state dynamic model over the nominal state (linearization point) can be obtained by Equations \ref{91a}-\ref{91e} in state space form as,
\begin{equation}
    \delta\dot{\mathbf{x}}_I = \mathbf{F}(\mathbf{x}_I)\delta\mathbf{x}_I+ \mathbf{G}(\mathbf{x}_I)\mathbf{w}
\end{equation}
where
\begin{conditions*}
\mathbf{F}(\mathbf{x}_I)\in\mathbb{R}^{15\times15} & Linearized IMU state Jacobian over nominal state $\mathbf{x}_I$\\
\mathbf{G}(\mathbf{x}_I)\in\mathbb{R}^{15\times12}  & IMU process noise Jacobian over nominal state $\mathbf{x}_I$
\end{conditions*}

Note that $\mathbf{w}=\left[\mathbf{n}_{g}^\intercal,\mathbf{n}_{a}^\intercal,\mathbf{n}_{\mathbf{b}_{g}}^\intercal,\mathbf{n}_{\mathbf{b}_{a}}^\intercal\right]^\intercal$ is the process noise of the dynamic, we assume the time-invariant process noise covariance matrix $\mathbf{Q}$ is known by sensor calibration. The IMU state error covariance $\mathbf{P}_I$ can be propagated through continuous-time Riccati differential equation as follow,
\[\dot{\mathbf{P}}_I=\mathbf{F}(\mathbf{x}_I)\mathbf{P}_I+\mathbf{P}_I\mathbf{F}^\intercal(\mathbf{x}_I)+\mathbf{G}(\mathbf{x}_I)\mathbf{Q}\mathbf{G}(\mathbf{x}_I)^\intercal\]
where
\begin{conditions*}
\mathbf{P}_I\in\mathbb{R}^{15\times15}& Continuous-time IMU state error covariance\\
\mathbf{Q}\in\mathbb{R}^{12\times12}& Continuous-time IMU process noise covariance matrix 
\end{conditions*}

However, obtaining the IMU error covariance in a new time step requires solving the matrix Riccati differential equation by techniques like matrix fraction decomposition. Alternatively, for simplicity, we can discretize the linearized error state dynamics, then propagate the error covariance through the discretized linearized error state dynamics in Equation \ref{aa}. By exact integration with the constant assumption of Jacobian $\mathbf{F}\left(\mathbf{x}_{I_k}\right)$ and $\mathbf{G}\left(\mathbf{x}_{I_k}\right)$ over the interval $[t_k,t_{k+1}]$, we obtain
\begin{equation}
\delta\mathbf{x}_{I_{k+1}}=e^{\mathbf{F}\left(\mathbf{x}_{I_{k}}\right)\left(t_{k+1}-t_{k}\right)}\delta\mathbf{x}_{I_{k}}+\int_{t_{k}}^{t_{k+1}}e^{\mathbf{F}\left(\mathbf{x}_{I_{k}}\right)\left(t_{k+1}-\tau\right)}\mathbf{G}\left(\mathbf{x}_{I_{k}}\right)\mathbf{w}\mathrm{d}\tau\label{aa}
\end{equation}
where
\begin{conditions*}
e^{\mathbf{F}\left(\mathbf{x}_{I_{k}}\right)\left(t_{k+1}-t_{k}\right)}\in\mathbb{R}^{15\times15} & Matrix exponential (state-transition matrix) over interval $[t_k,t_{k+1}]$. For simplicity, denote it as $\boldsymbol{\Phi}(t_{k+1},t_k)$
\end{conditions*}

Given the discretized error state dynamic, the error state covariance matrix is propagated,

\begin{equation}
\mathbf{P}_{I_{k+1}}=\boldsymbol{\Phi}(t_{k+1},t_k)\mathbf{P}_{I_{k}}\boldsymbol{\Phi}^\intercal(t_{k+1},t_k)+\mathbf{Q}_k
\end{equation}
where
\begin{conditions*}
\mathbf{P}_{I_{k}}\in\mathbb{R}^{15\times15} & IMU state error covariance in time step $k$\\
\mathbf{Q}_k\in\mathbb{R}^{15\times15}& Discrete-time process noise covariance matrix, derived by Equation \ref{cc}  
\end{conditions*}
\begin{equation}
    \mathbf{Q}_k=\int_{t_k}^{t_{k+1}} \boldsymbol{\Phi}\left(t_{k+1}, \tau\right) \mathbf{G}\left(\mathbf{x}_{I_{k}}\right)\mathbf{Q} \mathbf{G}^{\intercal}\left(\mathbf{x}_{I_{k}}\right) \boldsymbol{\Phi}^{\intercal}\left(t_{k+1}, \tau\right) \mathrm{d}\tau\label{cc}
\end{equation}

% \mathbf{P}_{I_{k+1}}^{-}\in\mathbb{R}^{15\times15} & A priori error state covariance in time step $k+1$\\
% \mathbf{P}_{I_{k}}^{+}\in\mathbb{R}^{15\times15} & A posteriori error state covariance in time step $k$\\
In conclusion, the noise-free IMU state dynamic and its linearized error state dynamic enable the propagation of the state estimate and uncertainty. After deriving the corresponding discretization models, the IMU propagation step in the filter-based framework and the IMU cost in the optimization-based framework can be obtained as follows with the notation $\|\boldsymbol{a}\|^2_{\mathbf{M}}=\boldsymbol{a}^\intercal\mathbf{M}^{-1}\boldsymbol{a}$,
\begin{table}[H]
\centering
\begin{tabular}{ccc}
 \textbf{Filter-Based Propagation} & {\quad}& \textbf{Optimization-Based IMU Cost}\\

         \\
    $\begin{aligned} % "[c]" is the default
    \mathbf{x}_{I_{k+1}}&=\mathbf{f}_{d}\left(\mathbf{x}_{I_{k}},\tilde{\mathbf{u}}_k,0\right)\\
    \mathbf{P}_{I_{k+1}}&=\boldsymbol{\Phi}(t_{k+1},t_k)\mathbf{P}_{I_{k}}\boldsymbol{\Phi}^\intercal(t_{k+1},t_k)+\mathbf{Q}_k
         \end{aligned}$  & {\quad}& $\begin{aligned} % "[c]" is the default
         \mathbf{C}_{I_k}=\sum_{I_k\in\mathcal{I}_k}\left\| \mathbf{x}_{I_{k+1}}\ominus\mathbf{f}_{d}\left(\mathbf{x}_{I_{k}},\tilde{\mathbf{u}}_k,0\right)\right\|^{2}_{\mathbf{P}_{I_{k+1}}}
         \end{aligned}$ \\
\end{tabular}
\end{table}

The optimization cost is in typical nonlinear least square form, which requires iterative optimization based on linearization with reasonable initialization. We will discuss this later in Section \ref{obm} and \ref{ci}.

% \noindent\begin{minipage}{.5\linewidth}
%  \begin{center}
%      \textbf{Filter-based framework}
%  \end{center}
% \begin{equation}
%   a = b + c.
% \end{equation}
% \end{minipage}%
% \begin{minipage}{.5\linewidth}
%  \begin{center}
%      \textbf{Optimization-based framework}
%  \end{center}
% \begin{equation}
%   d = e + f.
% \end{equation}
% \end{minipage}

% In general, without the help of exteroceptive sensors, IMU state estimate is propagated by integration of equation \ref{5a}-\ref{5e}. However, the uncertainty of IMU state estimations grows rapidly due to noise propagation through integration. In this case, complementary exteroceptive sensors like cameras can be used to update the estimation, more details are in section \ref{F_O}.

\subsection{Camera and its Measurement Model}
Camera is the classical exteroceptive sensor. It captures the visual information of the visible local regions from scene radiance to pixel brightness.
There are various cameras for vision-based motion estimation including monocular, stereo, RGB-D, fisheye, catadioptric, and event cameras. Monocular cameras have the simplest setup with only one camera, but it needs to take into account scale ambiguity. This scale vague can be resolved by stereo cameras using two-view geometry. RGB-D cameras have better depth estimation under textureless conditions with the help of an active infrared emitter added to stereo cameras. In terms of large field of view (FoV) cameras (fisheye and catadioptric cameras), Zhang $et$ $al.$ in \cite{RN729} found that they are more suitable in narrow and small environments, while smaller FoV cameras perform better in larger scale scenarios. In recent years, event cameras have gained much attention for their high frame rates and wide dynamic range by only capturing the brightness changes at pixel level\cite{RN323}. Regarding image capture modes of the camera, there are two distinct modes: Rolling Shutter and Global Shutter. Rolling shutter cameras read off image row by row, while global shutter cameras capture the whole picture simultaneously. In most VI-SLAM-related datasets, a global shutter camera is used by default, although it is more expensive than a rolling shutter camera. There are two major types of projection models for describing camera measurements: perspective and catadioptric projection\cite{RN303}. In this work, we focus on the classical pinhole perspective projection model. 
We assume data association (feature extraction and matching or tracking) is done in this article. First, transform the 3D landmark from the world frame to the camera frame. Note that we use the Euclidean parametrization of 3D landmarks for generality.

\begin{flalign}
&\quad\quad\text{\textbf{Transformation:}}&&{}_{C}\mathbf{L}=\mathbf{h}_\mathbf{T}\big(\mathbf{x}_I,{}_{W}\mathbf{L},(\mathbf{R}_{CI},{}_{C}\mathbf{t}_{CI})\big)\triangleq\mathbf{R}_{CI}\mathbf{R}_{IW}\left({}_{W}{\mathbf{L}}-{}_{W}\mathbf{t}_{WI}\right)+{}_{C}\mathbf{t}_{CI} &&
\end{flalign}
where 
\begin{conditions*}
{}_{W}\mathbf{L}=[x_w,y_w,z_w]^\intercal\in\mathbb{R}^3&Landmark in world frame, ${}_{W}\mathbf{t}_{WL}$\\
{}_{C}\mathbf{L}=[x_c,y_c,z_c]^\intercal\in\mathbb{R}^3&Landmark in camera frame, ${}_{C}\mathbf{t}_{CL}$\\
\mathbf{R}_{IW}\in \mathrm{SO}(3)&Rotation from world frame to IMU frame obtained from $\mathbf{x}_I$\\
{}_{I}\mathbf{t}_{IW}\in\mathbb{R}^{3}&Position of world frame wrt. IMU frame expressed in IMU frame obtained from $\mathbf{x}_I$\\
(\mathbf{R}_{CI},{}_{C}\mathbf{t}_{CI})\in \mathrm{SO}(3)\times\mathbb{R}^{3}& Camera-IMU extrinsic
\end{conditions*}

Then the 3D point in the camera frame is then projected into the 2D image frame.

\begin{flalign}
    &\quad\quad\text{\textbf{Projection:}}&&\mathbf{z}_p=\mathbf{h}_\mathbf{P}\left({}_{C}\mathbf{L}\right)\triangleq
\left[\begin{array}{l}
x_c/z_c \\
y_c/z_c  \\
\end{array}\right]&&
\end{flalign}
where
\begin{conditions*}
{\color{black}\mathbf{z}_p=\left[x_p,y_p\right]^\intercal\in\mathbb{R}^2}& Projected 2D points in image frame
\end{conditions*}

The projected 2D point is then distorted by the distortion function, 
\begin{flalign}
    &\quad\quad\text{\textbf{Distortion:}}&&\mathbf{z}_d=\mathbf{h}_\mathbf{D}\big(\mathbf{z}_p,(\mathbf{k},\mathbf{p})\big)
\triangleq\underbrace{\frac{1+k_{1} r^{2}+k_{2} r^{4}+k_{3} r^{6}}{1+k_{4} r^{2}+k_{5} r^{4}+k_{6} r^{6}}}_{\text {Radial Distortion }}\begin{bmatrix}
x_p \\
y_p \\
\end{bmatrix}+\underbrace{\begin{bmatrix}\begin{array}{l}
2 p_{1} x_p y_p+p_{2}\left(r^{2}+2 {x_p}^{2}\right) \\
2 p_{2} x_p y_p+p_{1}\left(r^{2}+2 {y_p}^{2}\right)
\end{array}\end{bmatrix}}_{\text {Tangential Distortion}}&&
\end{flalign}
where $r^{2}={x_p}^{2}+{y_p}^{2}$,
\begin{conditions*}
{\color{black}\mathbf{z}_d=\left[x_d,y_d\right]^\intercal\in\mathbb{R}^2}& Distorted 2D points in image frame\\
r\in\mathbb{R}& Radius from the origin of the image to projected point in meter\\
\mathbf{k}=[\mathrm{k}_1,\mathrm{k}_2,\mathrm{k}_3,\mathrm{k}_4,\mathrm{k}_5,\mathrm{k}_6]^\intercal\in\mathbb{R}^{6}& Radial distortion coefficients\\
\mathbf{p}=[\mathrm{p}_1,\mathrm{p}_2]^\intercal\in\mathbb{R}^{2}& Tangential distortion coefficients\\
\end{conditions*}

Finally, the distorted 2D points in the image frame will move the origin from the center to the top right corner and transfer the units from meters to pixels, as shown below.

\begin{flalign}
    &\quad\quad\text{\textbf{Origin shift and Unit change:}}&&\tilde{\mathbf{z}}=\mathbf{h}_\mathbf{K}\left(\mathbf{z}_d,\mathbf{K}\right)+\mathbf{v}\triangleq\begin{bmatrix}
    f_{x} & 0\\
0 & f_{y} 
\end{bmatrix}\begin{bmatrix}
x_d \\
y_d
\end{bmatrix}+ \begin{bmatrix}
    c_x\\
    c_y
\end{bmatrix}+\begin{bmatrix}
    n_u\\
    n_v
\end{bmatrix} &&
\end{flalign}
where
\[\mathbf{K}=\begin{bmatrix}
    f_{x} & 0 & c_{x} \\
0 & f_{y} & c_{y} \\
0 & 0 & 1
\end{bmatrix}\]
\begin{conditions*}
{\color{black}\tilde{\mathbf{z}}=\left[u,v\right]^\intercal\in\mathbb{R}^2}&Distorted camera measurement in image frame in pixel unit \\
{\color{black}c_x,c_y\in\mathbb{R}}&Principal point in origin shifted image frame in pixel unit\\
{\color{black}f_x,f_y\in\mathbb{R}}&Focal lengths in pixel\\
\mathbf{K}\in\mathbb{R}^{3\times3}& Camera intrinsics matrix\\
\mathbf{v}=[n_u,n_v]^\intercal\in\mathbb{R}^{2}&Image noise vector
\end{conditions*}

The camera-IMU extrinsic, camera intrinsic, and distortion coefficients are assumed to be given through camera calibration. In summary, a single 3D landmark can be observed by the camera frame in the camera measurement model in Equation \ref{eq1},

\begin{equation} \label{eq1}
\begin{split}
\tilde{\mathbf{z}}&=\mathbf{h}\big(\mathbf{x}_I,{}_{W}\mathbf{L},(\mathbf{R}_{CI},{}_{C}\mathbf{t}_{CI}),(\mathbf{k},\mathbf{p}),\mathbf{K}\big)+\mathbf{v}\\
    &\triangleq\mathbf{h}_\mathbf{K}\left(\mathbf{h}_\mathbf{D}\left(\mathbf{h}_\mathbf{P}\left(\mathbf{h}_\mathbf{T}\big(\mathbf{x}_I,{}_{W}\mathbf{L},(\mathbf{R}_{CI},{}_{C}\mathbf{t}_{CI})\big)\right),(\mathbf{k},\mathbf{p})\right),\mathbf{K}\right)+\mathbf{v}
\end{split}
\end{equation}

Since the image distortion can be removed in a preprocessing step, the undistorted camera measurement model can be obtained in Equation \ref{eq2},

\begin{equation} \label{eq2}
\begin{split}
\tilde{\mathbf{z}}&=\mathbf{h}\big(\mathbf{x}_I,{}_{W}\mathbf{L},(\mathbf{R}_{CI},{}_{C}\mathbf{t}_{CI}),\mathbf{K}\big)+\mathbf{v}\\
    &\triangleq\mathbf{h}_\mathbf{K}\left(\mathbf{h}_\mathbf{P}\left(\mathbf{h}_\mathbf{T}\big(\mathbf{x}_I,{}_{W}\mathbf{L},(\mathbf{R}_{CI},{}_{C}\mathbf{t}_{CI})\big)\right),\mathbf{K}\right)+\mathbf{v}
\end{split}
\end{equation}

Since $(\mathbf{R}_{CI},{}_{C}\mathbf{t}_{CI}),\mathbf{K}$ are assumed known constants and camera and IMU are time synchronized, the undistorted camera measurement model can be simplified as
\[\tilde{\mathbf{z}}=\mathbf{h}\left(\mathbf{x}_{I},{}_{W}\mathbf{L}\right)+\mathbf{v}\quad\xRightarrow[]{\text{indexing}}\quad
\tilde{\mathbf{z}}^{k}_j=\mathbf{h}\left(\mathbf{x}_{I_k},{}_{W}\mathbf{L}_j\right)+\mathbf{v}\]
where
\begin{conditions*}
\mathbf{x}_{I_k}\in\mathbb{R}^{3}\times\mathbb{S}^{3}\times\mathbb{R}^{3}\times\mathbb{R}^{3}\times\mathbb{R}^{3} & IMU state at time step $k$, where there is a time-synchronized camera frame token in this time step \\
{}_{W}\mathbf{L}_j\in\mathbb{R}^{3}& Landmark in world frame with a unique index $j$\\
\tilde{\mathbf{z}}^{k}_j\in\mathbb{R}^{2} & 2D feature measurement pixel location of single landmark $j$ token at camera frame at time step $k$ in image coordinate
\end{conditions*}

Note that 2D features in camera measurement contain both geometric and photometric information. In this case, there are two types of measurement errors: \textbf{geometric} and \textbf{photometric} differences, used as innovation terms for the update in the filter-based framework or as measurement residuals in the optimization-based framework. The geometric difference is normally called "reprojection error" in computer vision since it measures the difference between the landmark measurement and its prediction obtained by reprojecting the landmark prediction onto the predicted camera frame. Define the reprojection residual (measurement residual or innovation term) of one 2D feature as Equation \ref{99b},

\begin{flalign}
&\quad\quad\text{\textbf{Geometric Reprojection Residual:}}&&\mathbf{r}^{k}_j=\mathbf{r}_{\mathbf{g}}\left(\tilde{\mathbf{z}}^{k}_j, \mathbf{h}\left(\mathbf{x}_{I_k},{}_{W}\mathbf{L}_j\right)\right) \label{99b} &&
\end{flalign}
where 
\begin{conditions*}
\mathbf{r}^{k}_j\in\mathbb{R}\text{ or }\mathbb{R}^{2}\text{ or } \mathbb{R}^{3}& Reprojection residual of single landmark $j$ prediction and its measurement in camera frame at time step $k$ based on the geometric error metrics $\mathbf{r}_{\mathbf{g}}(\cdot)$. For example, it contains 2 dimensions in pixel location difference and 3 dimensions in unit bearing vector error.
\end{conditions*}

Based on the parametrization of 2D feature measurement, different error metrics can be used for parametrizing the reprojection error. One standard reprojection error is the pixel location difference (image plane error) on the image plane, Zhang $et$ $al.$ \cite{RN729} introduce a similar faster error metric on the unit plane. In order to present some typical error metrics, we first define back-projection $\boldsymbol{\pi}^{-1}(\cdot):\mathbb{R}^{2}\mapsto\mathbb{P}^{2}$ that recovers the bearing vector(up to scale) from the undistorted camera measurement as Equation \ref{n1},

\begin{flalign}
&\quad\quad\text{\textbf{Back-projection(up to scale):}}&&\mathbf{m}=\boldsymbol{\pi}^{-1}(\tilde{\mathbf{z}})\triangleq\mathbf{K}^{-1}\underline{\tilde{\mathbf{z}}} \label{n1} &&
\end{flalign}
where 
\begin{conditions*}
\mathbf{m}=[m_x,m_y,m_z]^\intercal\in\mathbb{P}^{2}& Bearing vector(up to scale) from measurement's camera frame\\
\underline{\tilde{\mathbf{z}}}=[\tilde{\mathbf{z}},1]^\intercal\in\mathbb{P}^{2}& Camera measurement in homogeneous form with symbol $\underline{\{\cdot\}}$
\end{conditions*}

Here are some of the commonly used error metrics for the geometric difference in Equation \ref{n1a}-\ref{n4a}.

\begin{subequations}
\begin{align}
    &\text{\textbf{Image Plane Error:}}&&\mathbf{r}_{\tilde{\mathbf{z}}}=\tilde{\mathbf{z}}-\mathbf{h}\left(\mathbf{x}_{I},{}_{W}\mathbf{L}\right)\label{n1a}\\
    &\text{\textbf{Unit Plane Error:}}&&\mathbf{r}_{\bar{\mathbf{m}}}=\begin{bmatrix}
        m_x/m_z\\
        m_y/m_z
    \end{bmatrix}-\begin{bmatrix}
        x_c/z_c\\
        y_c/z_c
    \end{bmatrix}\label{n2a}\\
    &\text{\textbf{Unit Bearing Vector Error:}}&&\mathbf{r}_{\hat{\mathbf{m}}}=\frac{\mathbf{m}}{\|\mathbf{m}\|}-\frac{{}_C\mathbf{L}}{\|{}_C\mathbf{L}\|}\label{n3a}\\
    &\text{\textbf{Bearing Angle Error:}}&&\mathrm{r}_{\theta}=\arccos(\frac{\mathbf{m}^\intercal{}_C\mathbf{L}}{\|\mathbf{m}\|\|{}_C\mathbf{L}\|})\label{n4a}
\end{align}
\end{subequations}
where 
\begin{conditions*}
\mathbf{r}_{\tilde{\mathbf{z}}}\in\mathbb{R}^{2}& Error on image plane\\
\mathbf{r}_{\bar{\mathbf{m}}}\in\mathbb{R}^{2}& Error on unit plane, where $\bar{\mathbf{m}}=\mathbf{m}/m_z$ is the bearing vector on unit plane\\
\mathbf{r}_{\hat{\mathbf{m}}}\in\mathbb{R}^{3}& Error between two unit bearing vectors, where $\hat{\mathbf{m}}=\mathbf{m}/\|\mathbf{m}\|$ is the unit bearing vector\\
\mathrm{r}_{\theta}\in\mathbb{R}& Angular error between two bearing vectors
\end{conditions*}

\begin{figure}[h]
    \centering
    \includegraphics[width=0.6\textwidth]{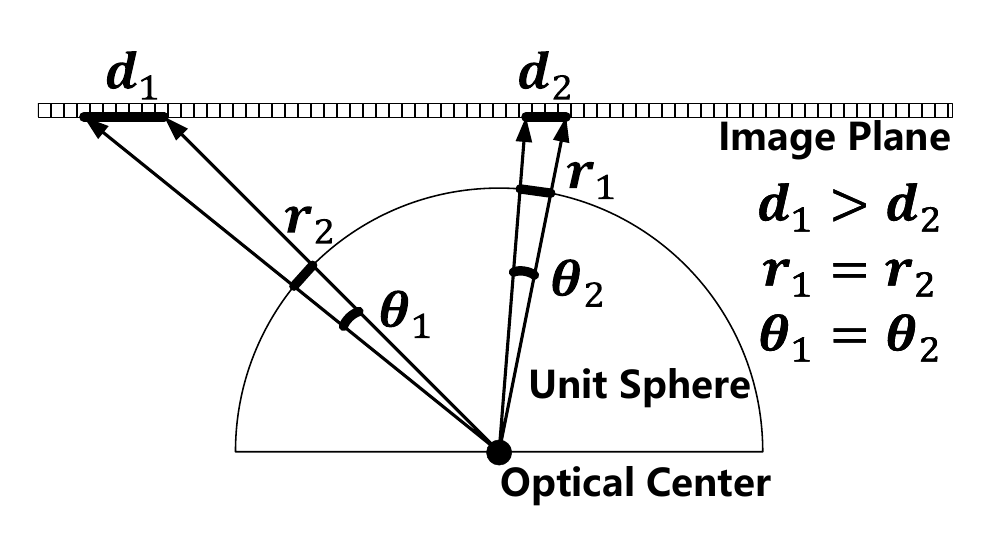}
    \caption{Reprojection Error on the image plane, unit sphere, and bearing angle in two different viewing angles with the same bearing angle difference, where $d_1$ and $d_2$ are the image plane error, $r_1$ and $r_2$ are unit bearing vector error, $\theta_1$ and $\theta_2$ are bearing angle error.}
    \label{fig:33}
\end{figure}

Essentially, the geometric information captured by the camera is  the bearing angle. For cameras with a large FoV, in-plane reprojection error becomes less sensitive at the edge of the plane where a large difference on the plane may correspond to a small bearing angle difference, as shown in Figure \ref{fig:33}. In this case, Zhang $et$ $al.$ \cite{RN729} suggest unit plane error metric for small FoVs cameras and unit bearing vector error for large FoVs cameras based on their efficiency and performance over FoVs. In this article, for generality, we adopt the standard image plane error as the geometric residual for further discussion.

Note that in the above geometric error metrics, the prediction of the 3D landmark can be parametrized as a bearing vector in the \textit{local} camera frame, which corresponds to the anchored point parametrization suggested by Solà $et$ $al.$ in \cite{RN7761}. In this case, we adopt the IDP parametrization\cite{RN686}\cite{RN1214} for 3D landmark prediction, which consists of a unit bearing vector $\hat{\mathbf{m}}$ and an inverse depth $\rho$ in the local frame. Here, we define the corresponding back-projection $\boldsymbol{\pi}^{-1}(\cdot):\mathbb{R}^{2}\times\mathbb{R}\mapsto\mathbb{R}^{3}$ that recovers the 3D landmark from IDP parametrization in Equation \ref{n11}.

\begin{flalign}
&\quad\quad\text{\textbf{Back-projection:}}&&{}_C \mathbf{L}=\boldsymbol{\pi}^{-1}(\tilde{\mathbf{z}},\rho)\triangleq \frac{\hat{\mathbf{m}}}{\rho} \label{n11} &&
\end{flalign}
where 
\begin{conditions*}
\hat{\mathbf{m}}=\mathbf{m}/\|\mathbf{m}\|\in\mathbb{S}^{2}& Unit bearing vector from camera frame origin to landmark\\
\rho\in\mathbb{R}& Inverse depth from camera frame origin to landmark
\end{conditions*}

This back-projected landmark in the camera frame can be inversely transformed into the world frame by the inverse transformation in Equation \ref{n111},
\begin{flalign}
&\quad\quad\text{\textbf{Inverse Transformation:}}&&{}_W \mathbf{L}=\mathbf{h}_\mathbf{T}^{-1}(\mathbf{x}_I,{}_{C}\mathbf{L})\triangleq \mathbf{R}_{IW}^\intercal\mathbf{R}_{CI}^\intercal\left({}_{C}{\mathbf{L}}-{}_{C}\mathbf{t}_{CI}\right)+{}_{W}\mathbf{t}_{WI} \label{n111} &&
\end{flalign}

Note that when evaluating the geometric difference between landmark measurement and prediction, the reprojection residual is constructed based on the 3D-to-2D correspondences. Compared to 2D-to-2D correspondences, 3D-to-2D correspondences include the estimation of 3D landmarks. According to Scaramuzza and Fraundorfer's tutorial \cite{RN303}, 2D-to-2D and 3D-to-2D methods are more accurate than 3D-to-3D methods. However, in the case of photometric difference, since the image intensity does not live in 3D space, there are only pair-wise 2D-to-2D correspondences (theoretically 2D-to-3D-to-2D) between two images in photometric residual. For representing photometric residual, we firstly define an image intensity function over a pixel location of camera frame at time $k$ as $\mathbf{I}_k(\cdot):\mathbb{R}^{2}\mapsto\mathbb{R}$. Since photometric residual is based on 2D-to-2D correspondences, we define this residual for single feature $j$ visible in an image pair at time step $k_1$ and $k_2$ as Equation \ref{999b},

\begin{flalign}
&\quad\quad\text{\textbf{Photometric Residual:}}&&\mathrm{r}_j^{k_1\veryshortarrow k_2}=\mathrm{r}_{\mathbf{p}}\left(\tilde{\mathbf{z}}^{k_1}_j,\rho^{k_1}_j,\mathbf{x}_{I_{k_1}},\mathbf{x}_{I_{k_2}}\right)\triangleq \mathbf{I}_{k_1}\left(\tilde{\mathbf{z}}^{k_1}_j\right)-\mathbf{I}_{k_2}\left(\mathbf{w}\left(\tilde{\mathbf{z}}^{k_1}_j\right)\right)\label{999b}&&
\end{flalign}
where wrap function $\mathbf{w}(\cdot):\mathbb{R}^{2}\mapsto\mathbb{R}^{2}$, that back-projecting a feature from $k_1$ frame to world frame then transforms and projects it onto $k_2$ frame, shown as follow,
\[\mathbf{w}\left(\tilde{\mathbf{z}}^{k_1}_j\right)=\mathbf{h}\Bigl(\mathbf{x}_{I_{k_2}},\mathbf{h}_\mathbf{T}^{-1}\bigl(\mathbf{x}_{I_{k_1}},\boldsymbol{\pi}^{-1}(\tilde{\mathbf{z}}^{k_1}_j,\rho^{k_1}_j)\bigl)\Bigl)\]
\begin{conditions*}
\mathrm{r}_j^{k_1\veryshortarrow k_2}\in\mathbb{R}& Photometric difference of feature $j$ between its intensity measurement on $k_1$ frame and its predicted location's intensity on $k_2$ frame based on photometric error metrics $\mathrm{r}_{\mathbf{p}}(\cdot)$.\\
\tilde{\mathbf{z}}^{k_1}_j\in\mathbb{R}^{2}& 2D feature measurement pixel location of single landmark $j$ token at camera frame at time step $k_1$ in image coordinate\\
\rho^{k_1}_j\in\mathbb{R} & Inverse depth of 3D landmark corresponding to feature $j$ on $k_1$ frame\\
\mathbf{x}_{I_{k_1}},\mathbf{x}_{I_{k_2}}& IMU states at time step $k_1$ and $k_2$
\end{conditions*}

Note that in Equation \ref{999b}, the photometric residual is computed over a pair of pixels, which is commonly used in dense (DTAM\cite{RN89}, DVO\cite{RN83}\cite{RN897}) and semi-dense (LSD-SLAM\cite{RN77}) feature frameworks for ease of computation. However, in sparse (VI-DSO\cite{RN78}\cite{RN456}, ROVIO\cite{RN399}, SVO\cite{RN78}\cite{RN551}) feature frameworks, this photometric residual is computed in a pair of pixel patches that contain the weighted average intensity difference of neighboring pixels centered on the feature pixel. In DSO\cite{RN76}, Engel $et$ $al.$ evaluate the accuracy and efficiency of nine different pixel patch patterns. They also use a more precise photometric camera model that considers exposure time, camera response function, and vignetting effect, since the photometric errors are sensitive to varying brightness.
For simplicity, we adopt pixel-wise instead of patch-wise photometric residual for ease of notation.

In order to update the state and error covariance in the filter-based framework or construct the visual cost (bundle adjustment (BA) in computer vision) on the optimization-based framework, we first define the measurement Jacobian matrix $\mathbf{H}_k$ over the linearization point $\mathcal{X}_k$. This measurement Jacobian matrix can be represented in rows form, where one rows is corresponding to a single feature 2D measurement. It also can be divided into columns, where left columns are derivatives over IMU states $\frac{\partial \mathbf{h}}{\partial \mathcal{I}}$ and right columns are derivatives over landmarks $\frac{\partial \mathbf{h}}{\partial \mathcal{L}}$ for all measurements.

\begin{flalign}
&\quad\quad\text{\textbf{Measurement Jacobian:}}&&\mathbf{H}_k=\begin{bmatrix}
    \frac{\partial \mathbf{h}_1}{\partial \mathcal{X}}\\
    \vdots\\
    \frac{\partial \mathbf{h}_j}{\partial \mathcal{X}}\\
    \vdots\\
    \frac{\partial \mathbf{h}_m}{\partial \mathcal{X}}\\
\end{bmatrix}=\begin{bmatrix}
    \frac{\partial \mathbf{h}}{\partial \mathcal{I}}&\frac{\partial \mathbf{h}}{\partial \mathcal{L}}
\end{bmatrix},\quad\frac{\partial \mathbf{h}_j}{\partial \mathcal{X}}=\begin{bmatrix}
    \frac{\partial \mathbf{h}_j}{\partial \mathcal{I}}&\frac{\partial \mathbf{h}_j}{\partial \mathcal{L}}
\end{bmatrix}\label{09b}&&
\end{flalign}
where $\frac{\partial \mathbf{h}_j}{\partial \mathcal{X}}=\left.\frac{\partial \mathbf{h}_j}{\partial \mathcal{X}}\right|_{\mathcal{X}=\mathcal{X}_k}$ is the Jacobian corresponding to feature $j$ 2D measurement. The Jacobian of each measurement can be divided into columns, which include derivatives over IMU states $\frac{\partial \mathbf{h}_j}{\partial \mathcal{I}}$ and derivatives over landmarks $\frac{\partial \mathbf{h}_j}{\partial \mathcal{L}}$.

The visual update step in the filter-based framework and visual cost in the optimization-based framework can be obtained as follows, where $\|\cdot\|_{\gamma}$ is a robust penalty function ($L_1$, Huber, Cauchy, etc.). $\mathbf{R}$ denotes the visual measurement noise covariance.

\begin{table}[H]
\centering
\begin{tabular}{cc}
 \textbf{Filter-Based Upate} & \textbf{Optimization-Based Visual Cost}\\

         \\
    $\begin{aligned} % "[c]" is the default
    \mathbf{K}&=\mathbf{P}\mathbf{H}_k^{\intercal}\left(\mathbf{H}_k\mathbf{P}\mathbf{H}_k^{\intercal}+\mathbf{R}\right)^{-1}\\
    \mathbf{x}_{I_{k}}&=\mathbf{x}_{I_{k}}+\mathbf{K}\mathbf{r}\\
    \mathbf{P}_{I_{k}}&=\left(\mathbf{I}-\mathbf{K} \mathbf{H}_k\right) \mathbf{P}_{I_{k}}\left(\mathbf{I}-\mathbf{K} \mathbf{H}_k\right)^{\intercal}+\mathbf{K} \mathbf{R}\mathbf{K}^{\intercal}
         \end{aligned}$  & $\begin{aligned} % "[c]" is the default
         \mathbf{C}_{V_k}=\begin{cases}
             \sum\limits_{k\in\mathcal{I}_k}\sum\limits_{\mathbf{L}\in\mathcal{L}_k}\sum\limits_{j\in\mathrm{obs}(\mathbf{L})}\left\| \mathbf{r}_{\mathbf{g}}\left(\tilde{\mathbf{z}}^{k}_j, \mathbf{h}\left(\mathbf{x}_{I_k},{}_{W}\mathbf{L}_j\right)\right)\right\|^{2}_{\mathbf{R}^{k}_{j}}\\
             \sum\limits_{k\in\mathcal{I}_k}\sum\limits_{\mathbf{L}\in\mathcal{L}_k}\sum\limits_{j\in\mathrm{obs}(\mathbf{L})}\sum\limits_{i\in\mathbf{N}(k)}\left\| \mathrm{r}_{\mathbf{p}}\left(\tilde{\mathbf{z}}^{k}_j, \rho^{k}_j,\mathbf{x}_{I_{k}},\mathbf{x}_{I_{i}}\right)\right\|_{\gamma}
         \end{cases}
         \end{aligned}$ \\
\end{tabular}
\end{table}
where in the filter-based update, $\mathbf{r}$ is the visual measurement residuals (innovation) from the geometric $\mathbf{r}_{\mathbf{g}}(\cdot)$ or photometric $\mathrm{r}_{\mathbf{p}}(\cdot)$ difference. In optimization-based visual cost, $k\in\mathcal{I}_k$ denotes the frame index $k$ in IMU frame set ${I}_k$, $\mathbf{L}\in\mathcal{L}_k$ denotes landmark $\mathbf{L}$ in landmark set $\mathcal{L}_k$, $j\in\mathrm{obs}(\mathbf{L})$ denotes the feature index $j$ from observable landmark $\mathbf{L}$ in IMU frame $k$ corresponded camera frame. $i\in\mathbf{N}(k)$ denotes the neighborhood frame $i$ around frame $k$.

% Augmented state, structureless framework.

\subsection{Factor Graph representation of VIN}\label{fac}

\begin{figure}[H]
     \centering
     \begin{subfigure}[b]{0.48\textwidth}
         \centering
         \includegraphics[width=\textwidth]{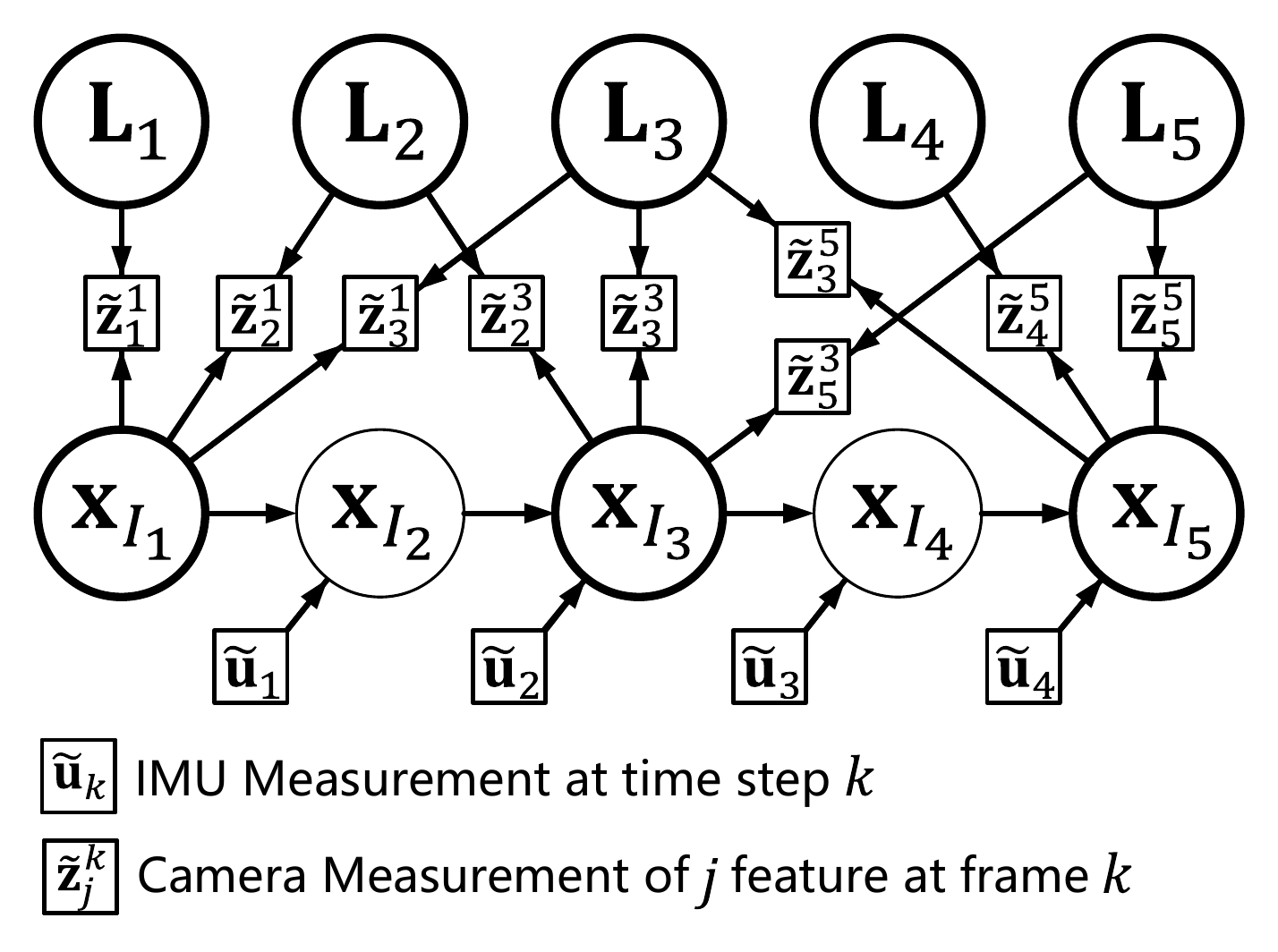}
         \caption{Bayesian Network}
         \label{fig:g2}
     \end{subfigure}
     % \hfill
     \begin{subfigure}[b]{0.51\textwidth}
         \centering
         \includegraphics[width=\textwidth]{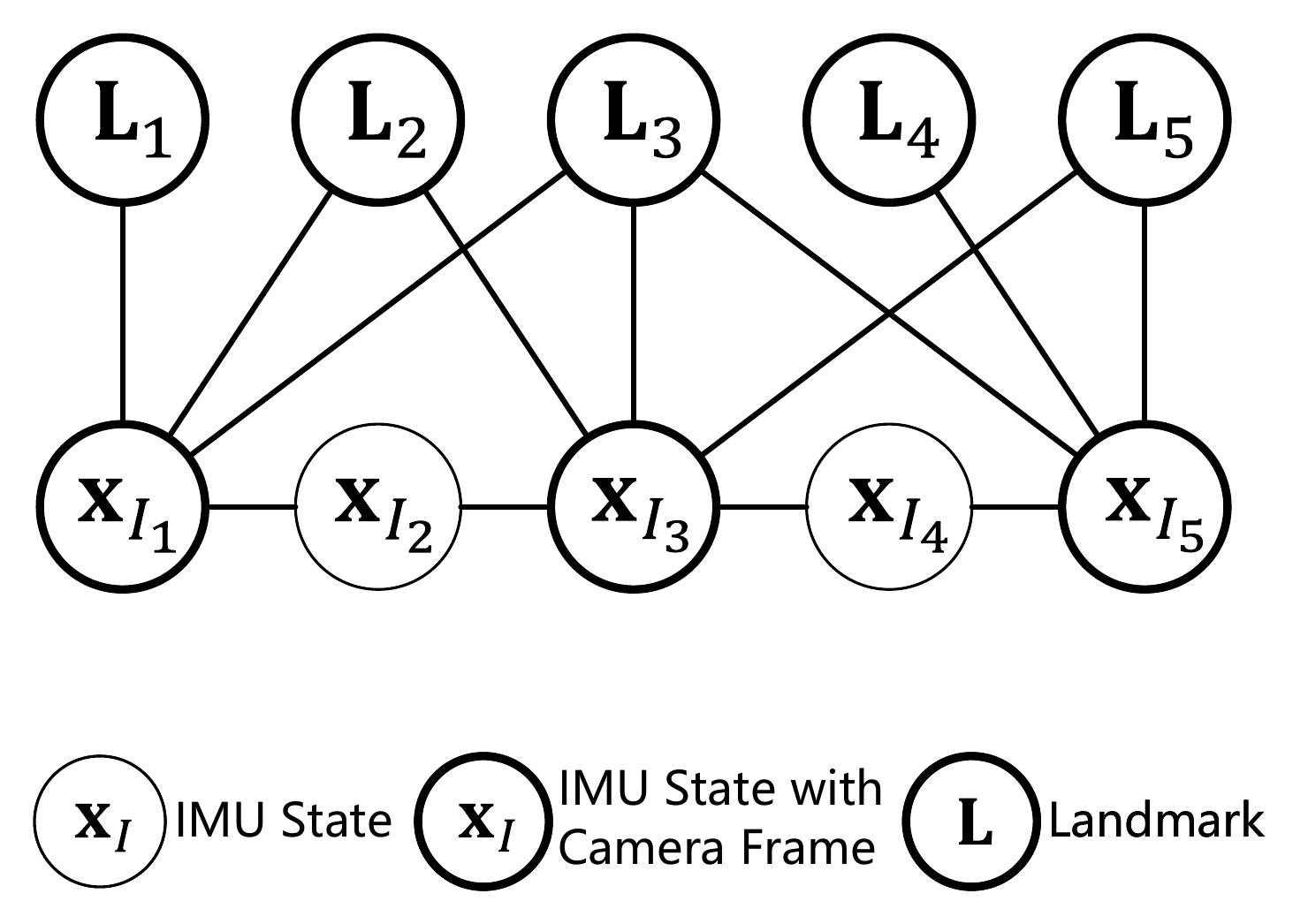}
         \caption{Markov Random Field}
         \label{fig:g1}
     \end{subfigure}
        \caption{Probabilistic Graphical Models in VIN}
        \label{g12}
\end{figure}

The evolution of visual-inertial navigation can be naturally represented in probabilistic graphical models in terms of a directed acyclic graph (Bayesian Network) or an undirected graph (Markov Random Field), as shown in Figure \ref{g12}. In the Bayesian network shown in Figure \ref{fig:g2}, nodes clearly indicate three types of quantities: to be estimated states, observed measurements, and controllable actions, while in VIN, control inputs are replaced by IMU measurements. Arrows in the Bayesian network clearly indicate causal relationships between nodes, in which the IMU state is propagated through IMU measurements, and camera measurements depends on both the state of IMU and landmarks. In Markov Random Fields shown in Figure \ref{fig:g1}, the dependencies of nodes are clearly presented through undirected links. The sampling rate of the IMU measurements is much faster than the camera measurements, and in this case, since we assume IMU and camera measurements are time-synchronized, there will be several IMU states between two camera-synchronized IMU states (For simplicity, there is only one IMU state between the two camera-synced IMU states in Figure \ref{g12}). Under the Gaussian uncertainty assumption, VIN can be represented in Gaussian Markov Random Fields\cite{RN1174}, where the linkage and correlation between nodes can be characterized by the information matrix. The information matrix of the graphical model in VIN preserves certain sparsity. The nonzero entries of the information matrix indicate the links between nodes, and the magnitude of entries indicates the "strength" of the link(correlation between nodes). The sparsity of the information matrix directly relates to the computational efficiency of VIN\cite{RN355}. Optimizing the VIN problem based on its graph structure is important for the efficiency of graph-based optimization, i.e. reasonably reducing the number of nodes (marginalization) and links (sparsification)\cite{RN483}\cite{RN630}. For example, marginalizing the pass nodes in the filter (MSCKF\cite{RN397}, ROVIO\cite{Bloesch9}\cite{RN399}, OpenVINS\cite{RN398}) and fixed-lag smoother (OKVIS\cite{RN309}, VINS-Mono\cite{RN91}, VI-DSO\cite{RN456}, BASALT\cite{RN474}), and removing the weak links for sparsification (SEIF\cite{RN355}\cite{RN1167}). Marginalization and sparsification will eventually affect the pattern of the information matrix, we refer readers to Eustice's thesis\cite{RN1170} for an in-depth understanding of inference in the information form. 

\begin{figure}[h]
    \centering
    \includegraphics[width=0.6\textwidth]{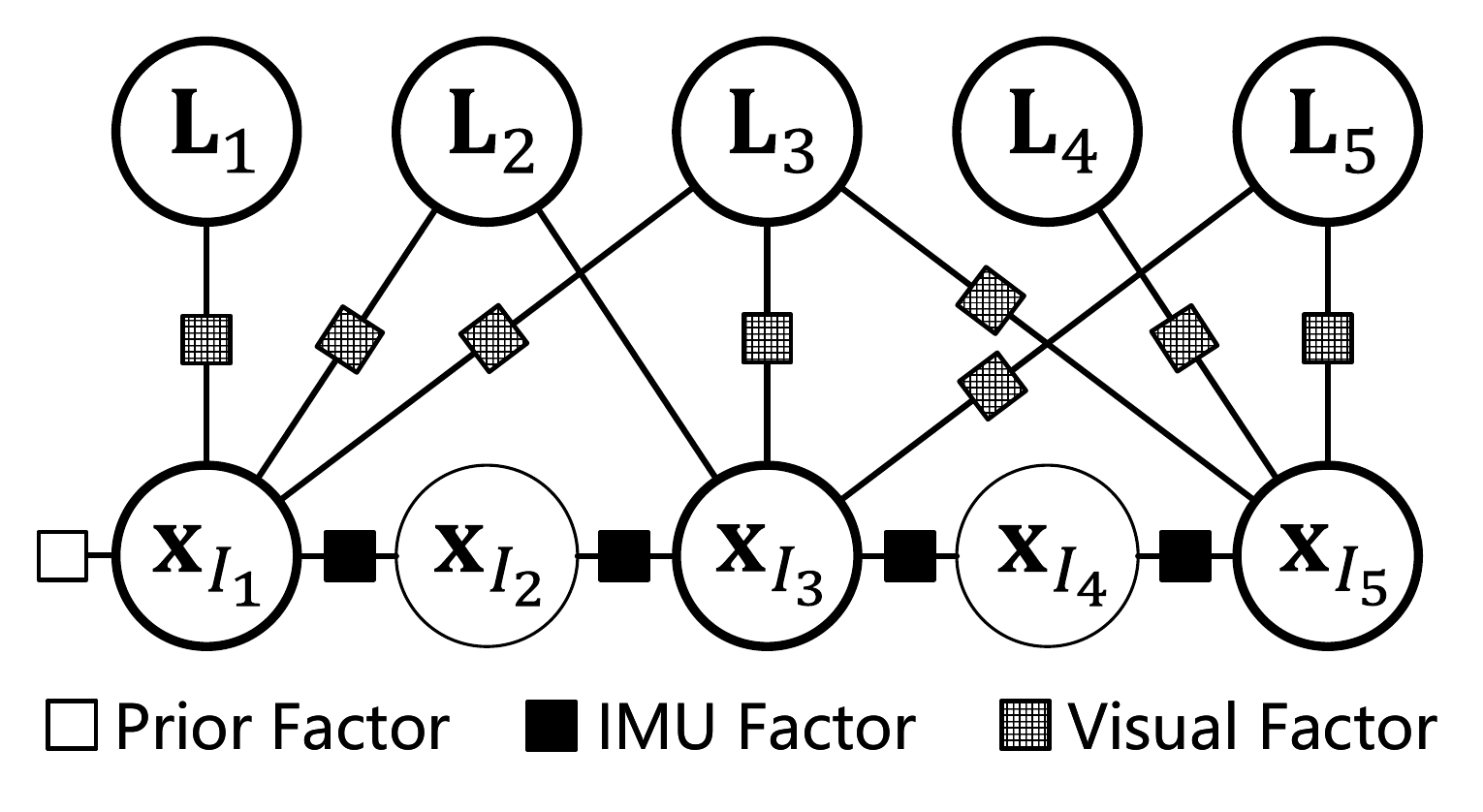}
    \caption{Factor Graph in VIN}
    \label{fig:g3}
\end{figure}

Factor graph is a bipartite graph originally operated by sum-product algorithm (belief propagation algorithm)\cite{RN311}\cite{RN1180}. It contains two types of nodes: variables (unknown states) and factors (known functions). Compared to Bayesian Network and Markov Random Field, factor graphs present relationships (factors) between variables. VIN's factor graph has three types of factors: an optional prior factor, IMU factors, and visual factors. The prior factor is a prior distribution over the  IMU states, by including this factor, the minimization of optimization cost in VIN can be viewed as a maximum a posteriori (MAP) estimate. IMU factors are characterized by IMU dynamic model with IMU measurements and their noises and visual factors are characterized by the camera measurement model with camera measurements and their noises. Note that in this section the graph includes all states (full batch optimization) at each time step and is incremented over time. The prior distribution describes the uncertainty of the initial state, whereas, in fixed-lag smoother, it contains prior information from marginalization. The full factor graph describing VIN is shown in Figure \ref{fig:g3}.

\begin{figure}[h]
    \centering
    \includegraphics[width=0.7\textwidth]{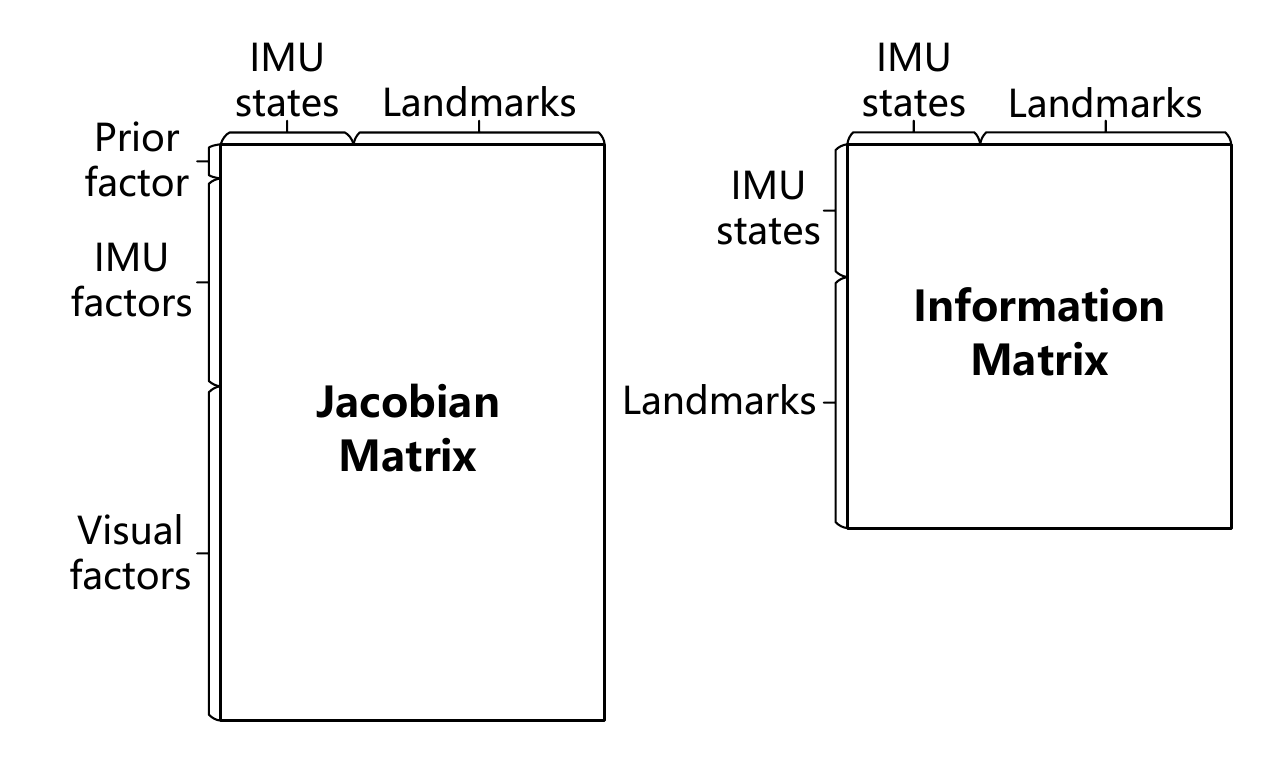}
    \caption{Jacobian and Hessian(Information Matrix) in VIN}
    \label{fig:ji}
\end{figure}

As the relationship between the Gaussian Markov Random field and information matrix, the factor graph is naturally related to the Jacobian matrix in VIN where rows in Jacobian show factors that indicate the relation between variables. IMU factors and visual factors are binary factors, where IMU factors link two IMU states, and visual factors link one IMU state and one landmark respectively. The prior factor is a uni-nary factor for the initial state in batch optimization or n-nary factors in fixed-lag smoother depending on marginalization.
A conceptual diagram representing the relationship between rows and columns in the Jacobian and Hessian (information matrix) is shown in Figure \ref{fig:ji}.

The factor graph of VIN can be formulated as an optimization cost function in terms of factor terms. The full cost of VIN in batch optimization includes three costs: Prior cost, IMU cost, and Visual cost shown in Equations \ref{a}-\ref{c}, where the prior cost of the initial states assumes Gaussian prior distribution with mean $\mathbf{x}_{p}$ and covariance $\mathbf{P}_{p}$. 

\begin{subequations}
\begin{align}
    &\text{\textbf{Prior Cost:}}&&\mathbf{C}_{p}=\|\mathbf{x}_{I_{0}}\ominus\mathbf{x}_{p}\|^2_{\mathbf{P}_{p}}\label{a}\\
    &\text{\textbf{IMU Cost:}}&&\mathbf{C}_{I_k}=\sum_{I_k\in\mathcal{I}_k}\left\| \mathbf{x}_{I_{k+1}}\ominus\mathbf{f}_{d}\left(\mathbf{x}_{I_{k}},\tilde{\mathbf{u}}_k,0\right)\right\|^{2}_{\mathbf{P}_{I_{k+1}}}
    \label{b}\\
    &\text{\textbf{Visual Cost:}}&& \mathbf{C}_{V_k}=\begin{cases}
             \sum\limits_{k\in\mathcal{I}_k}\sum\limits_{\mathbf{L}\in\mathcal{L}_k}\sum\limits_{j\in\mathrm{obs}(\mathbf{L})}\left\| \mathbf{r}_{\mathbf{g}}\left(\tilde{\mathbf{z}}^{k}_j, \mathbf{h}\left(\mathbf{x}_{I_k},{}_{W}\mathbf{L}_j\right)\right)\right\|^{2}_{\mathbf{R}^{k}_{j}}& \text{Geometric}\\
             \sum\limits_{k\in\mathcal{I}_k}\sum\limits_{\mathbf{L}\in\mathcal{L}_k}\sum\limits_{j\in\mathrm{obs}(\mathbf{L})}\sum\limits_{i\in\mathbf{N}(k)}\left\| \mathrm{r}_{\mathbf{p}}\left(\tilde{\mathbf{z}}^{k}_j, \rho^{k}_j,\mathbf{x}_{I_{k}},\mathbf{x}_{I_{i}}\right)\right\|_{\gamma} & \text{Photometric}
         \end{cases}
    \label{c}
\end{align}
\end{subequations}

The full cost can be written in Equation \ref{d}. By minimizing the cost, the optimal states and their uncertainties estimation can be obtained, with Gaussian assumption, the optimal states and uncertainties are presented in means and covariances (or information matrix) respectively.

\begin{equation}
    \mathbf{C}(\mathcal{X}_k)=\mathbf{C}_{p}+\mathbf{C}_{I_k}+\mathbf{C}_{V_k}\label{d}
\end{equation}

\section{State Estimation Methods in VIN}\label{abcc}
In the preview section, we introduce the relevant quantities (states of interest and support quantities) and models (IMU and camera models) in VIN and their graph-based representations. In this section, we discuss the existing methodologies for VIN state estimation. In this tutorial, we mainly focus on tightly-coupled visual-inertial fusion. There are two main tightly-coupled visual-inertial state estimation schemes: filter-based and optimization-based. As such, we briefly classify these state estimation methods in terms of the width of the state horizon.

\begin{itemize}
    \item \textbf{Filter (one-step recursive estimation):}
    
    It normally estimates the latest state given the latest measurement. It boosts efficiency but loses accuracy due to the marginalization of all past information. The loss of accuracy is also due to the accumulation of linearization errors\cite{RN400}.
    % Filter-based estimation methods include EKF, UKF, and particle filters.
    \item \textbf{Fixed-lag smoother (moving horizon estimator or sliding window filter):}
    
    It estimates the recent states given the recent measurements. It balances the speed and precision of estimation by changing the width of the moving horizon. Note that fixed-lag smoother is normally based on optimization.
    
    \item \textbf{Batch estimator (full horizon estimation or full smoother):}
    
    It estimates all states given all measurements using nonlinear optimization. It gradually becomes computationally intractable with continuously increasing states.
    % , but this can be alleviated by incrementally solving the nonlinear estimation problem instead of repeating batch steps\cite{RN476}, or 
\end{itemize}

\begin{figure}[H]
     \centering
     \begin{subfigure}[b]{0.45\textwidth}
         \centering
         \includegraphics[width=\textwidth]{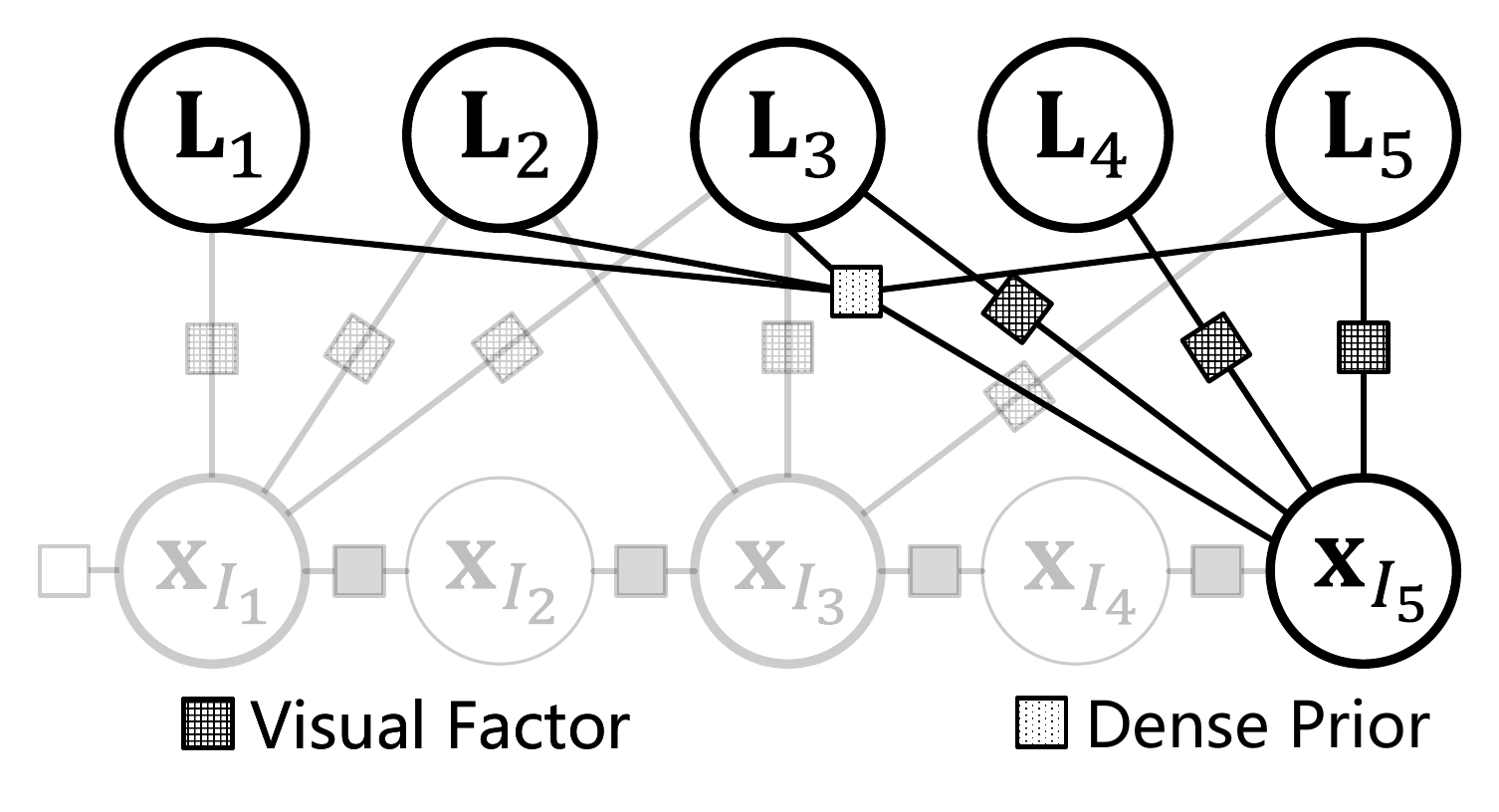}
         \caption{Filter}
         \label{fig:t2}
     \end{subfigure}
     \hfill
     \begin{subfigure}[b]{0.45\textwidth}
         \centering
         \includegraphics[width=\textwidth]{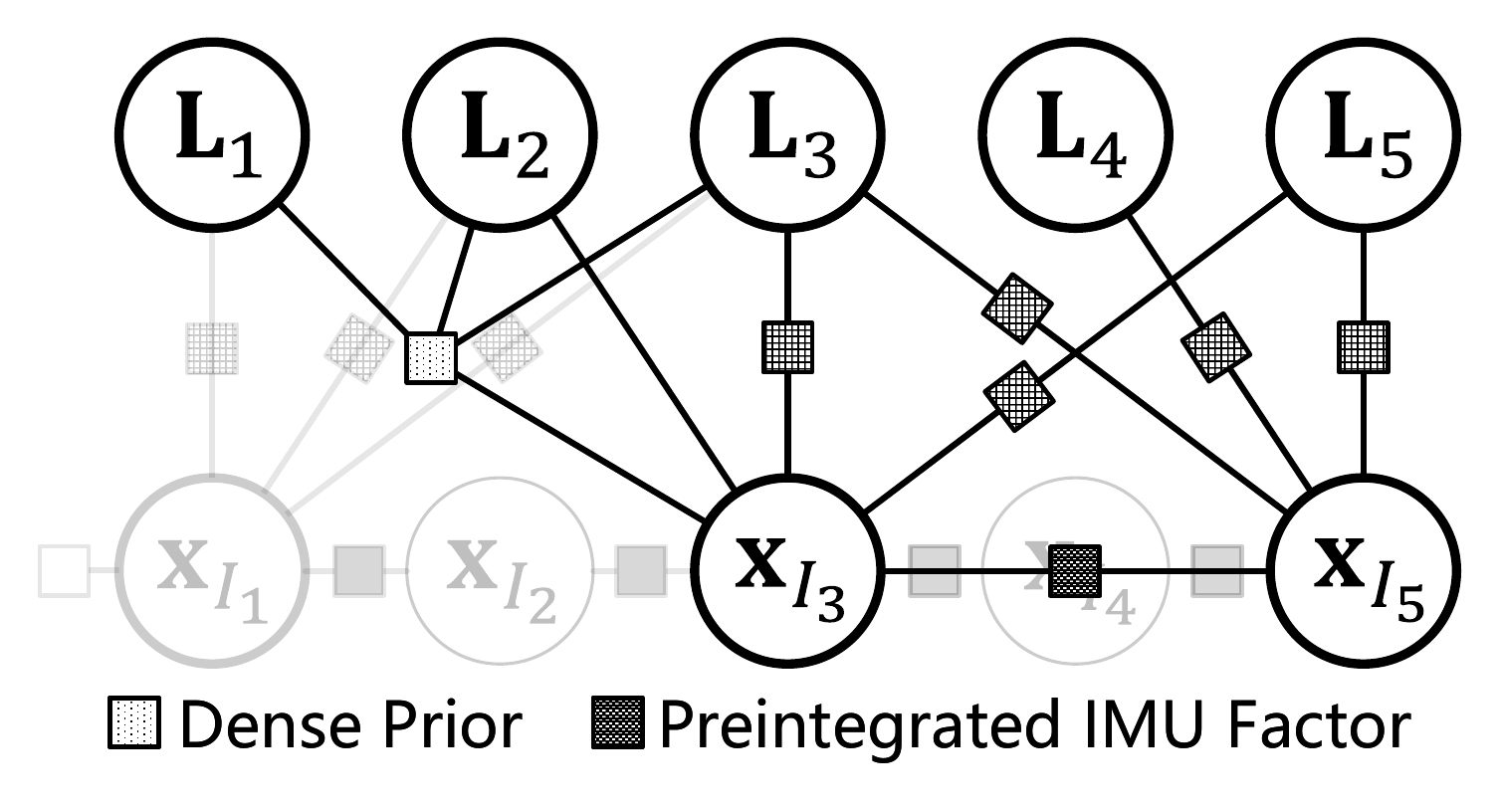}
         \caption{Fixed-lag Smoother}
         \label{fig:t1}
     \end{subfigure}
        \caption{Filter and Smoother in VIN}
        \label{t12}
\end{figure}

The factor graphs of the filter and the fixed-lag smoother are shown in Figure \ref{fig:t2} and \ref{fig:t1} respectively. The factor graph of the batch estimator is shown in the previous section in Figure \ref{fig:g3}. Note that due to the marginalization of past states, a dense prior factor will be introduced in both filter and fixed-lag smoother. This will increase the density of the information matrix, and eventually reduce the efficiency of the estimation. In this case, to reduce the computational burden from the "fill-in" in the information matrix caused by the densely connected prior factor, certain sparsification methods are designed to balance the efficiency and accuracy (either "break" the links or "drop" the nodes).

\subsection{Filter-based Methods}

\begin{figure}[h]
    \centering
    \includegraphics[width=0.8\textwidth]{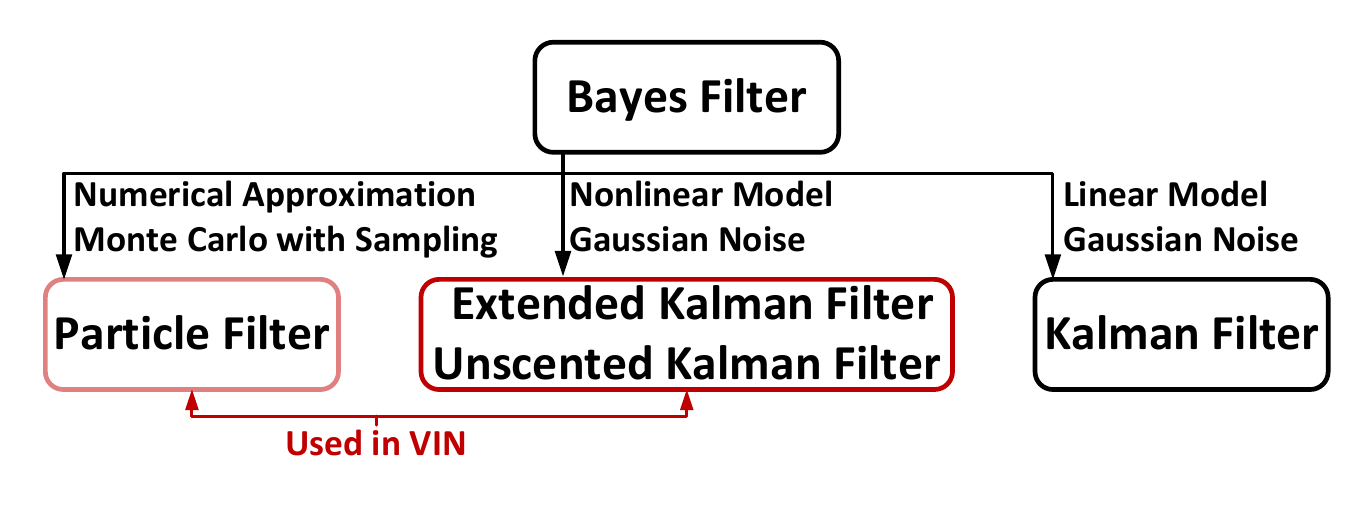}
    \caption{Filter-based Methods in VIN}
    \label{bayes}
\end{figure}

All filter-based methods in VIN are built on the foundation of Bayes filters. Bayes filters provide a unified framework for probabilistic state estimation\cite{RN586}. In the Bayes filter, the state information is represented by the probability distribution (probability density function), and the evolution of the state information can be characterized as two stages: Propagation and Update. In the propagation phase, the state loses information due to propagating the state through the dynamic model, while in the update phase, the state gains information by obtaining measurements from the measurement model. However, parameterizing state information as probability distributions is computationally intractable. Numerical approximations or assumptions should be made to deploy Bayesian concepts into actual implementations. The typical numerical approximation of the Bayes filter is the particle filter based on Monte Carlo simulation and importance sampling techniques. Also, under the linear Gaussian assumption, the Bayesian filter becomes a Kalman filter, which is the optimal state estimator under this assumption. However, in VIN, both the IMU dynamics and camera measurement models are nonlinear. In this case, keeping the reasonable Gaussian noises assumption, the Extended Kalman Filter (EKF) and Unscented Kalman Filter (UKF) are often used in VIN to cope with nonlinear models. In Figure \ref{bayes}, We briefly illustrate the relationship between these filters.

The particle filter is capable of working under the nonlinear model with non-Gaussian noise. However, since particle filters are built based on importance sampling techniques, a large number of particles (sampling points) are required to characterize the distributions of high-dimensional states (curse of dimensionality). In this case, the Rao-Blackwellisation technique is used to marginalize out some of the variables\cite{RN366}. Rao-Blackwellised particle filters (RBPF) have been successfully applied in many visual-based navigation works, such as FastSLAM\cite{RN364}, FastSLAM2\cite{RN365}, and its improved versions\cite{RN459}\cite{RN79}\cite{RN435} with inverse depth landmark initialization, fewer particles, and better particle weight computation respectively. Although the above-mentioned RBPFs work well in the 2D environment, they are not computationally effective compared to UKFs and EKFs methods with high-dimensional states, such as VIN with motion in three dimensions, more dynamic states (velocity), and sensor biases\cite{RN902}. Both UKFs and EKFs are designed to overcome nonlinearities in the models, with Gaussian noise assumptions, and they have similar theoretical complexities. EKFs require Jacobian over linearization points, while UKFs need to calculate the sigma points in each iteration. Compared to the analytical Jacobian matrix, the sigma points require repeated calculations and are more computationally intensive, but UKFs and EKFs have similar performance in SLAM\cite{RN903}. In this case, filter-based methods in VIN mostly concentrate on the EKF and its dual-form extended information filter (EIF) because of their efficiency. EKFs are initially implemented in visual-only SLAM, including MonoSLAM\cite{RN387}\cite{RN387} that consider a constant velocity motion model for active feature searching, and robocentric frameworks\cite{RN962}\cite{RN1206} that change the reference frame from world to local for improving the consistency of EKFs-based SLAM. In order to reduce computational burdens while keeping the information, sparsification methods are introduced including sparse extended information filters (SEIF)\cite{RN355}\cite{RN1167} that neglecting or conservatively eliminating the distant landmarks\cite{RN1167} in the information form. 

The observability and consistency of the proposed estimator are always of-interest in filter-based methods. In \cite{RN837}, an analysis of the observability of non-linear 2D world-centric SLAM is proposed, which indicates that the direct use of linear observability tools yields inconsistency. This problem is further investigated in EKF-based SLAM\cite{RN838} by Huang $et$ $al.$ and first-estimates Jacobian EKF (FEJ-EKF)\cite{RN831} is proposed to improve the consistency and reduce the linearization error, where Jacobians are computed using the first available estimate for each state variable. FEJ-EKF is further improved by observability-constrained EKF (OC-EKF)\cite{RN624}, which selects the linearization points that preserve observability while minimizing the error between the linearization point and true state. The concept of OC-EKF was then applied to 3D VIN by Hesch $et$ $al.$'s work observability constrained VINS (OC-VINS) \cite{RN626}\cite{RN568}\cite{RN829}.

One of the most well-known filter-based methods in VIN is the multi-state constraint Kalman filter (MSCKF)\cite{RN397} introduced by Mourikis and Roumeliotis in 2007. MSCKF is a structureless EKF-based method that marginalizes all landmarks by nullspace operation. Bloesch $et$ $al.$ propose ROVIO\cite{Bloesch9}\cite{RN399}, an EFK and iterated extended Kalman filter(IEKF) based method that utilizes the photometric innovation instead of typical geometric innovation for state update. ROVIO includes the dynamic of a fixed number of stationary landmarks for robust estimation. 
While conventional EKFs operate in vector spaces, some states of VIN lie in Lie groups. In this case, Brossard $et$ $al.$ propose a UKF on Lie groups\cite{RN436}\cite{RN434} based on invariant Kalman filter(I-KF) theory\cite{RN477}. Solà presents a dedicated tutorial on error-state extended Kalman filter(ES-EKF)\cite{RN396} that enables uncertainty propagation of the quaternion Lie group under error-state dynamics. 

% Overall, the filter-based methods focus more on the efficiency of the estimation. Additionally, filter-based methods in VIN more concentrate on the estimation of ego-motion rather than environment mapping. 

\subsection{Optimization-based Methods}\label{obm}

Compared to EKF-based methods, most optimization-based methods are also based on linearization operations but are able to iteratively re-linearize at new linearization points and repeat until the optimization converges. In \cite{RN506}, Bell and Cathey prove the equivalency between IEKF and Gauss-Newton optimization at filter update. This relinearization process in optimization reduces the linearization error. The optimization-based methods are also called graph-based methods\cite{RN656} in SLAM. As indicated in Section \ref{fac}, VIN has a well-forming graphical interpretation in factor graphs. The optimization problem in VIN can be formulated as a summation of cost functions over the factors as shown in Equations \ref{a}-\ref{c} including prior, IMU, and visual cost. Over the past decade, optimization-based methods have experienced tremendous development due to the maturity of optimization tools for state estimation including Google Ceres\cite{RN1131}, g2o\cite{RN306}, GTSAM\cite{RN451}, etc. There are many optimization-based masterpieces in VIN are introduced during this period, including OKVIS\cite{RN309}, SVO+GTSAM\cite{RN505}\cite{RN400}, ORB-VIO\cite{RN405}, IS-VIO\cite{RN484}, VINS-Mono\cite{RN91}, VI-DSO\cite{RN456}, Kimera\cite{RN530}, BASALT\cite{RN474}, DM-VIO\cite{RN455}, etc. To balance efficiency and accuracy, these fixed-lag smoother-based methods maintain a subset of the full state compared to batch estimators. OKVIS\cite{RN309} introduces a recent keyframe-based graph structure as a moving horizon in the fixed-lag smoother framework. VINS-Mono\cite{RN91} and SVO+GTSAM\cite{RN505}\cite{RN400} remove IMU states between consecutive camera keyframes by using IMU preintegrated factors. ORB-VIO\cite{RN405} uses the covisibility graph that includes the past camera frames sharing enough common features with the current keyframes. IS-VIO\cite{RN484} introduces a specific sparsification method to sparsify dense priors while keeping information loss to a minimum. Kimera\cite{RN530} supports metric-semantic mapping with dense mesh. BASALT\cite{RN474} adopts the sparsification methods proposed in NFR\cite{RN630} that maintain the information measured by Kullback–Leibler divergence after marginalizing nodes. The above-mentioned methods are indirect methods that minimize the geometric error in optimization, where VI-DSO\cite{RN456} and DM-VIO\cite{RN455} are direct methods based on photometric error. They both include the scale and gravity direction in the tight-coupled optimization while utilizing dynamic and delayed marginalization respectively.

For long-term large-scale mapping, batch estimators with loop closure are commonly used in VIN. In this scenario, the computational complexity increases unboundedly as the graph grows. Kaess $et$ $al.$ propose incremental nonlinear optimization methods: iSAM\cite{RN561} and iSAM2\cite{RN476}, that avoiding repeated batch processing steps. However, the computational complexity depends not only on the increasing scale of the graph but also on the connection density of the graph. In this case, nodes marginalization(or removal) and edges sparsification are introduced including generic node removal (GLC)\cite{RN483} and nonlinear factor recovery (NFR)\cite{RN630}.

\subsection{On-manifold Operation}\label{onm}

In VIN, the motion of the pose is normally represented by rigid body transformation in special Euclidean group $\mathrm{SE}(3)$, where rotation has various parametrizations with different properties discussed in Section \ref{s32}. All of these parameterizations related to pose transformation are Lie groups, which is a smooth manifold. However, perturbations in Lie groups are not simple addition and subtraction operations like in vector spaces. Fortunately, all elements in the Lie group can be mapped into a corresponding tangent space (Lie algebra) in its vector form. In this way, optimization tools that typically operate on vector spaces can be used for on-manifold optimization. The bijective mapping functions between Lie group $\mathcal{M}$ and Lie algebra in vector space $\mathbb{R}^{m}$ are Exponential mapping function $\mathrm{Exp}(\cdot):\mathbb{R}^{m}\mapsto\mathcal{M}$ and Logarithm mapping function $\mathrm{Log}(\cdot):\mathcal{M}\mapsto\mathbb{R}^{m}$. In the previous section, we denote generic plus and minus operations $\oplus$ and $\ominus$, we specify this operation in vector space and Lie groups respectively as followed.
\begin{align*}
    &&&\text{\textbf{Vector Space}}&&\text{\textbf{Lie group and Lie algebra}}\\
    &\text{\textbf{PLus} }(\oplus)&&\mathbf{y}=\mathbf{x}+\delta\mathbf{x}&&\mathcal{Y}=\mathcal{X}\mathrm{Exp}\left({}^{\mathcal{X}}\boldsymbol{\tau}\right)\\
    &\text{\textbf{Minus} }(\ominus)&& \delta\mathbf{x}=\mathbf{y}-\mathbf{x}&&{}^{\mathcal{X}}\boldsymbol{\tau}= \mathrm{Log}\left(\mathcal{X}^{-1}\mathcal{Y}\right)
\end{align*}
where we adopt the right-hand convention
\begin{conditions*}
\mathbf{x},\mathbf{y}\in\mathbb{R}^m& Quantities in vector space\\
\delta\mathbf{x}\in\mathbb{R}^m& Perturbation around $\mathbf{x}$ in vector space\\
\mathcal{X},\mathcal{Y}\in\mathcal{M}& Quantities in Lie group\\
{}^{\mathcal{X}}\boldsymbol{\tau}\in\mathbb{R}^m& Perturbation over $\mathcal{X}$ (right-hand convention) in Lie algebra with vector form
\end{conditions*}

The back-and-forth mapping (Exponential and Logarithm) between the manifold and tangent space enables the iterative optimization algorithms on manifolds, such as gradient descent, Gauss-Newton, and Levenberg-Marquardt methods, etc. Solà $et$ $al.$ provide a detailed tutorial on Lie theory for robotics state estimation\cite{RN243}, Forster $et$ $al.$ also provide a comprehensive formulation of on-manifold operations for IMU preintegration\cite{RN505}\cite{RN400}. In addition, these on-manifold operations are important for uncertainty propagation on Lie groups. By assuming Gaussian noise on the tangent space of the manifold, Barfoot and Furgale\cite{RN533} give an accurate characterization of uncertainty propagation over $\mathrm{SE}(3)$ with left-hand convention. Mangelson $et$ $al.$\cite{RN1156} extend the work of Barfoot and Furgale by considering jointly correlated poses.

\subsection{Calibration and Initialization in VIN}\label{ci}
Calibration in VIN is to estimate the model-related quantities (mostly time-invariant) discussed in Section \ref{s32}. Since two types of sensory modalities are used in VIN, the calibration involves self-calibration and inter-calibration over visual and inertial sensors. Furthermore, these quantities can be calibrated during estimation (online calibration) or beforehand (offline calibration). Camera self-calibration mainly involves the estimation of the camera intrinsic\cite{RN633}\cite{RN634} including the camera's principal point, focal length, and distortion parameters. For stereo camera setup, stereo rectification\cite{RN852} is also needed which makes the epipolar lines horizontal with proper scale for directly capturing the image disparity. Among the direct methods relying on the photometric differences of feature pixels, a more accurate photometric camera model is needed that takes into account the time-varying auto-exposure time, camera response function, and vignetting-induced attenuation factors\cite{RN301}\cite{RN456}\cite{Bloesch9}. IMU calibration normally requires the assistance of exteroceptive sensors (magnetometers or cameras). In IMU calibration, the most concerned quantities are the gyroscope and accelerometer biases. However, these two biases are time-varying quantities as random walk processes, and in this case, both biases are included in the state vector in VIN. The noise parameters in IMU can be obtained by using Allan standard deviation\cite{RN748}\cite{RN745}\cite{RN827}. For low-cost consumer-grade IMU, the effects of axis misalignment, scale factor errors, and linear acceleration on the gyroscope should be also considered\cite{RN408}\cite{RN1159}. IMU-camera calibration is to estimate the geometric and temporal difference between two types of sensors. The geometric difference is normally called IMU-camera extrinsic, which includes the displacement and orientation between two sensors. The temporal difference is the time offset between camera and IMU measurements caused by their different latency and sampling rate. IMU-camera calibration can be solved offline by kalibr\cite{RN406}\cite{RN408}. The IMU-camera extrinsic is calibrated online in many VIN works including OKVIS\cite{RN309}, ROVIO\cite{Bloesch9}\cite{RN399}, VINS-Mono\cite{RN91}, and OpenVINS\cite{RN398} in filter-based or optimization-based frameworks. The time offset between IMU and camera measurements can also be estimated online in \cite{RN407}\cite{RN92} or avoided by hardware synchronization\cite{RN419}.

The initialization is critical for the nonlinear least square optimization in fixed-lag smoother and batch estimator in VIN. Proper initialization can prevent the optimization from converging to false local minima due to nonlinearity and non-convexity. Although, Olson $et$ $al.$\cite{RN482} have shown that  stochastic gradient descent (SGD) can be used for poor initial estimates. In optimization-based visual-only and visual-inertial navigation, initialization consists of the majority of the workload since optimization toolboxes(Ceres, g2o, GTSAM, etc) are used for final code implementation. The initialization from the visual-only navigation relies on the classical multiple-view geometry in computer vision. In quantities required to be initialized in tight-coupled optimization-based VIN contains active states (the pose of camera frames corresponding IMU frames, velocity, gyroscope and accelerometer biases, and 3D landmarks position), scale, and gravity direction. For the monocular camera, the relative 2-view camera pose can be estimated by the commonly used five-point algorithm\cite{RN383} by Nister or eight-point algorithm\cite{RN982} by Longuet-Higgins, or by decomposing a homography matrix if viewing planar scenes\cite{RN1137}. After obtaining the relative pose (rotation and up-to-scale translation), all features observed in these two frames can be triangulated to estimate their 3D positions up to scale. Then the absolute pose of the camera frame that observes these 3D landmarks knowing their 3D positions can be estimated by the perspective-n-point (PnP) methods like P3P\cite{RN636}\cite{RN637} and EPnP\cite{RN347}, etc. Kneip and Furgale provide a unified software library OpenGV\cite{RN519} for relative and absolute camera pose estimation. Dong-Si and Mourikis\cite{RN1003}\cite{RN1001} present closed-form solutions for VIN while considering the number of features and frames with certain trajectories needed for possible solutions. Martinelli\cite{RN411} derives intuitive closed-form solutions that investigate in detail the number of distinct solutions with different numbers of features and frames with different motions under the biased and unbiased case. In ORB-VIO\cite{RN405}, Mur-Artal and Tardos propose a novel IMU initialization method that first estimates gyroscope bias by consecutive keyframes visual odometry then approximates the scale and gravity direction given the preintegration over at least four keyframes, after this accelerometer bias is estimated and scale and gravity direction are refined in a similar manner, eventually velocity can be estimated given two consecutive position estimates, gravity vector estimate, and the preintegrated factors between them. In VINS-Mono\cite{RN91}, Qin $et$ $al.$ presents an efficient loosely coupled initialization procedure that ignores accelerometer bias, which is difficult to observe because most of the magnitude of the acceleration is due to gravity. Campos $et$ $al.$\cite{RN948} introduces an outstanding inertial-only optimization for visual-inertial initialization that jointly optimizes all the IMU-related variables and the scale factor in one step using MAP estimation.

\section{Performance Evaluation and Improvement in VIN}
The performance of VIN can be evaluated in terms of accuracy, efficiency, and robustness. Accuracy and efficiency can be assessed quantitatively, while robustness assessment is more qualitative based on illumination changes, motion blur, and low-texture scenes. Accuracy metrics in VIN include absolute trajectory error (ATE) and relative pose error (RPE) that evaluate the geometric error over the whole trajectory or over segmented sub-trajectory respectively\cite{RN315}\cite{RN318}\cite{RN376}. When considering the uncertainty of the estimate, the accuracy can be evaluated in terms of consistency by normalized estimation error squared (NEES), which can be tested only in simulations\cite{RN1089}. Efficiency metrics in VIN are normally evaluated in terms of computing resource usage and average processing time instead of floating point operations per second (FLOPs) for precise computational complexity measurement which is hard to compute depending on the convergence of iterative optimization and randomness induced by RANSAC. Delmerico and Scaramuzza present a detailed comparison\cite{RN394} of monocular VIO in terms of accuracy, latency, and computational constraints over popular VIO pipelines. We list some of the public-available datasets for VIN performance evaluation in Table \ref{table:my_label}.

\begin{table}[H]
    \centering
    \begin{tabular}{c|c|c|c|c|c}
     \textbf{Dataset} & \textbf{Environment} & \textbf{Platform} & \textbf{Groudtruth}& \textbf{Year} & \textbf{Ref.}\\
     \hline
     \hline
        NTU VIRAL & Outdoor & UAV & Total Station & 2022 & \cite{RN1235}\\
     \hline
     Newer College & Outdoor & Handheld & LiDAR Scan Prior Map & 2021 & \cite{RN325}\\
     \hline
     MADMAX & Outdoor & Rover&  RTK-GNSS & 2021 & \cite{RN1234}\\
     \hline
     Hilti & \specialcell{Indoor\\Outdoor} & Handheld & \specialcell{Motion Capture System\\Total Station}& 2021 &\cite{RN333}\\
     \hline
     4Seasons & Outdoor & Car & RTK-GNSS & 2020 & \cite{RN316}\\
     \hline
    UZH-FPV Drone & \specialcell{Indoor\\Outdoor} & UAV & Total Station & 2019 & \cite{RN377}\\
    \hline
    KAIST Urban & Outdoor & Car & \specialcell{RTK-GNSS\\SLAM} & 2019 & \cite{RN1233}\\
    \hline
    TUM VI & \specialcell{Indoor\\Outdoor} & Handheld & Motion Capture System & 2018 & \cite{RN319}\\
    \hline
    Canoe VI & Outdoor River & USV & GPS/INS& 2018 & \cite{RN349}\\
    \hline
    Oxford Car & Outdoor & Car & GPS/INS & 2017 & \cite{RN340}\\
    \hline
    NCLT & Outdoor & Rover & \specialcell{RTK-GNSS\\SLAM} & 2016 & \cite{RN1236}\\
    \hline
    EuRoC & Indoor & UAV & \specialcell{Motion Capture System \\Total Station} &2016 &\cite{RN314}\\
    \hline
    KITTI & Outdoor & Car & RTK-GNSS/INS & 2012 & \cite{RN315} \\
    \hline
    \end{tabular}
    \caption{VIN Datasets}
    \label{table:my_label}
\end{table}

\subsection{Accuracy, Efficiency, and Robustness Improvements}

In this section, we summarize efforts to improve the accuracy, efficiency, and robustness of existing filter-based and optimization-based frameworks. Since EKF-based methods are more efficient than UKF-based methods while having similar accuracy performance\cite{RN903}, we focus on analyzing EKF-based methods. EKF-based and optimization-based methods are normally based on linearization and Gaussian assumptions. In EKF-based frameworks, linearization error is the main cause of inaccuracy. In this case, inverse depth landmark parametrization is presented\cite{RN1223}\cite{RN686}\cite{RN1214}, which reduces the linearization error of the measurement model in low-parallax scenes. Robot-centric map joining\cite{RN1206}\cite{RN962} also improves the linearization of the model by the robot-centered representation that bounds the uncertainty. This linearization error can also be reduced by carefully selecting linearization points\cite{RN831}\cite{RN626} or iterative update linearization points in IEKF\cite{RN399}. However, these filter-based methods continuously marginalize past states, which leads to a densely connected prior factor with landmarks as shown in Figure \ref{fig:t2}. Strasdat $et$ $al.$\cite{RN1036}\cite{RN388} indicate that the computational cost of filter-based methods scales poorly with the number of landmarks. They conclude that the number of landmarks involved is the key factor for increasing the accuracy of visual SLAM, while in VIN, Bloesch $et$ $al.$\cite{RN399} point out that the quality of tracked landmarks is more important than the quantity because the IMU provides a good motion prior. Optimization-based methods utilize the iterative optimization methods that re-linearize the model in a new linearization point in each iteration, this mechanism naturally reduces the linearization error. These smoothing methods also maintain the past information that will be marginalized in the filter-based methods. For long-term navigation, loop closure by pose graph optimization and covisibility graph\cite{RN295}\cite{RN405} increase the accuracy while maintaining the efficiency.

The efficiency of VIN can be improved by reducing nodes and factors (edges in MRF) or incremental solutions. Nodes can be marginalized or removed, and resulting dense connections caused by fill-in in marginalization can be re-assigned by different topologies\cite{RN630} or simply breaking the weak link\cite{RN355}\cite{RN1167}. In VIN, the past visual nodes (camera frame with corresponding IMU frame) are marginalized in filter and fixed-lag smoother, which leads to a dense prior factor. In filter-based works, sparsification that drops the weak link is used\cite{RN355}\cite{RN1167}, whereas in fixed-lag smoother, specific sparsification(IS-VIO\cite{RN484}) that drops connections between landmarks and velocity and biases nodes or general sparsifications that enable different Markov blanket approximation including tree, subgraph and cliquey subgraph topologies while keeping the information loss to a minimum(NRF\cite{RN630}) are used for increasing the efficiency. In keyframes-based works including OKVIS\cite{RN309}, VINS-Mono\cite{RN91}, and SVO+GTSAM\cite{RN505}\cite{RN400}, the visual factors are simply dropped in non-keyframe visual nodes, and the IMU nodes are preintegrated as a preintegration factor between consecutive keyframes. In some extreme cases, all landmarks are marginalized out using nullspace operation\cite{RN397} or using smart factors\cite{RN1073}\cite{RN505}\cite{RN400}. Compared to feature-based indirect methods, direct methods are faster because they skip the data association stage avoiding the expensive feature extraction, matching, and outlier removal processes\cite{RN78}\cite{RN456}.

In real-world scenes, visual conditions involving dynamic objects, motion blur, low texture, and illumination changes often occur. In VIN, the combination of visual and inertial sensing modalities naturally robustifies the estimation since visual information avoids the drift caused by biased dead reckoning in inertial integration and IMU provides motion hints for visual sensing. The robustness of VIN can be further improved by alleviating the effect caused by misleading visual information, including outlier rejection and inlier selection or robust M-estimators. The outlier rejection is commonly achieved by RANSAC\cite{RN782} or Mahalanobis distance test over the landmark innovation and its covariance prediction\cite{RN397}\cite{Bloesch9}\cite{RN399}. By involving the motion model as constraints, fewer points are needed in RANSAC motion estimation procedures, from normally five\cite{RN383} or eight points\cite{RN982} for six-degree motion, to two points in 2D relative motion\cite{RN1012}, to one point in 2D motion considering Ackermann steering model\cite{RN384}\cite{RN960}\cite{RN385}. Additionally, the motion prior provided by IMU can be used to robustly track the static features in a highly dynamic environment\cite{RN993} with aggressive camera motion\cite{RN1023}. In terms of inlier selection, Zhao and Vela present an active feature selection and matching algorithm\cite{RN1243}\cite{RN1242} that reduces computational cost while maintaining the accuracy and robustness of pose tracking. The robust M-estimators are commonly used in VIN for reducing the effects caused by unrejected outliers. Mactavish and Barfoot provide a comparison\cite{RN1011} of different robust cost functions in visual navigation. Yang $et$ $al.$\cite{RN1199} introduce a robust penalty function with a control parameter for graduated non-convexity, they claim that the proposed approach can be a valid replacement for RANSAC with better performance in many spatial perception problems. In specific VIN works, ROVIO\cite{Bloesch9}\cite{RN399} robustifies its filtering estimation by considering the dynamics of the static landmarks. VINS-Mono\cite{RN91} presents a robust initialization procedure to deal with the gyro bias. Failure detection and recovery procedures are also important for the overall system robustness, as presented in typical direct methods (DSO\cite{RN76}, etc) and indirect methods (ORB-SLAM\cite{RN87}, VINS-Mono, etc). DSO detects the failure motion by examining the current motion's RMSE and attempts to recover by trying up to 27 different small rotations in different directions. ORB-SLAM detects the failure motion by examining the solutions provided by decomposing a homography or fundamental matrix and repeats the initialization process for recovery if not a clear winner solution is provided. VINS-Mono detects the failure by examining the number of features, continuity of position or rotation in consecutive frames, and change in gyro bias estimate. It repeats the initialization like ORB-SLAM for failure recovery.

\section{Learning Era in VIN}

Both the filter-based and optimization-based approaches discussed above are typical model-based methods built on models of IMU dynamics and camera measurements. Due to the vast development of deep learning, data-driven methods gradually challenge(or replace) the classical model-based methods in many areas, especially in the domain of natural language processing and computer vision. The successful data-driven applications related to visual and inertial navigation involve single image depth estimation (SIDE), deep visual odometry, deep inertial odometry, novel view synthesis (NVS), semantic SLAM, feature detection and matching, etc.
% For dense mapping, photometric-based direct methods are commonly used, where geometric information of feature pixels is parametrized as pixel location with depth toward the corresponding 3D landmark. In this case, estimating the depth of each pixel in images are important for dense mapping. 
Single image depth estimation\cite{RN610}\cite{RN720}\cite{RN721} provided a learning-based solution for feature depth estimation which can be used for monocular visual odometry and dense mapping. For deep visual odometry, Yang $et$ $al.$\cite{RN470} introduce D3VO that predicts monocular depth, photometric uncertainty, and relative camera pose in CNN-based network architectures. Koestler $et$ $al.$ present TANDEM\cite{RN469}, a real-time monocular tracking and dense mapping framework, which combines DSO\cite{RN76} and dense depth maps rendered from the global truncated signed distance function (TSDF) model to achieve visual odometry. Ummenhofer $et$ $al.$ introduce a supervised CNNs-based framework DeMoN\cite{RN464} for depth and pose estimation of two frames, where the network is based on encoder-decoder pairs with an iterative loop structure. For deep inertial odometry, Chen $et$ $al.$ propose IONet\cite{RN817}, which is the first end-to-end learning framework that takes raw IMU measurements and outputs 2D inertial odometry trajectories. Yan $et$ $al.$ provide RoNIN\cite{RN818} dataset and propose three deep network architectures based on ResNet, LSTM, and TCN for data-driven inertial navigation. Brossard $et$ $al.$ propose the AI-IMU\cite{RN437}, which adopts the filter-based framework for IMU state estimation and uses a deep neural network to dynamically adapt the noise parameters. Liu $et$ $al.$ introduce TLIO\cite{RN800} which incorporates learning-based displacement distribution estimation into EKF-based inertial odometry. Buchanan $et$ $al.$ introduce a learning-based IMU bias predictions\cite{RN1215} using two commonly used sequential networks: LSTMs and Transformers. 
In the case of novel view synthesis, Mildenhall $et$ $al.$ introduce neural radiance field (NeRF) \cite{RN1252}, a state-of-the-art view synthesis method using a fully connected deep network and principles from classical ray tracing.

Learning-based methods can also exact the semantic information in visual measurements.
Landmarks in the environment not only contain geometric information but also semantic category information, jointly utilizing the semantic and geometric can achieve more robust and informative landmarks distinguishment. Xiao $et$ $al.$ proposed Dynamic-SLAM\cite{RN907}, which added semantic segmentation to distinguish static and dynamic objects under the ORB-SLAM framework. Doherty $et$ $al.$ \cite{RN504} propose a robust semantic SLAM with probabilistic data association. Rosinol $et$ $al.$ provide an open-source C++ library Kimera\cite{RN530} that enables real-time VIN with 3D mesh reconstruction and semantic labeling.

In terms of feature extraction and matching, DeTone $et$ $al.$ propose a CNN-based feature detection and description algorithm SuperPoint\cite{RN297}. Sarlin $et$ $al.$ introduce SuperGlue\cite{RN794}, a learning-based feature matching based on graph neural networks. The data-driven approach achieves excellent performance across many components of the vision and initial navigation pipelines. However, there is still room for the complete end-to-end learning-based VIN. We refer the reader to paper \cite{RN854} for considering the limitations and potentials of the learning-based methods in robotics perception. Chen $et$ $al.$ provide a comprehensive survey\cite{RN492} for data-driven visual and inertial localization and mapping. 

\section{Conclusions}

Visual inertial navigation fuses the information provided by the camera and IMU sensor to obtain navigation-related geometric information. The complementary property and low-cost lightweight characteristic of these two sensory modalities make them popular in many navigation applications. VIN is a typical state estimation problem, in this article, we clearly define the relevant quantities in VIN and their parametrization and symbolization. Furthermore, the IMU dynamic and camera measurement models are also presented while considering IMU dynamic propagation with preintegration and its linearized discretized error state dynamic, and camera model with geometric and photometric residuals. In model-based methods including filter and optimization, VIN can be straightforwardly visualized from a factor graph perspective. The performance of VIN is continuously improved using graph-based optimization methods. The data-driven methods have revolutionized many aspects of the visual part of VIN, these learning-based methods provide alternatives with respect to the classical model-based methods including the end-to-end learning methods and hybrid methods that combine data-driven and model-based methods. Overall, This article hopes to provide a comprehensive overview of the VIN in terms of its relevant quantity presentation, model formulation, and possible methodologies with certain improvements.

% \printbibliography

\bibliography{reference.bib}

\end{document}